\documentclass[type=submission]{article}
\usepackage{ijcai26}

\usepackage{times}
\usepackage{soul}
\usepackage{url}
\usepackage[hidelinks]{hyperref}
\usepackage[utf8]{inputenc}
\usepackage[small]{caption}
\usepackage{graphicx}
\usepackage{amsmath}
\usepackage{multirow}
\usepackage{amsmath, amsfonts, amssymb, amsthm}
\usepackage{booktabs}
\usepackage{algorithm}
\usepackage{algorithmic}
\usepackage[switch]{lineno}

\usepackage{xcolor}

\newtheorem{proposition}{Proposition}

\usepackage[acronym]{glossaries}
\newacronym{MOO}{MOO}{Multi-Objective Optimization}
\newacronym{MOP}{MOP}{Multi-objective Optimization Problem}
\newacronym{MOPs}{MOPs}{Multi-objective Optimization Problems}
\newacronym{MOEA}{MOEA}{Multi-Objective Evolutionary Algorithm}
\newacronym{MOEAs}{MOEAs}{Multi-Objective Evolutionary Algorithms}

\newacronym{PS}{PS}{Pareto Set}
\newacronym{PF}{PF}{Pareto Front}
\newacronym{HV}{HV}{Hypervolume}
\newacronym{RMSE}{RMSE}{Root Mean Squared Error}
\newacronym{IGD+}{IGD+}{Inverted Generational Distance Plus}
\newacronym{CICP}{CICP}{Confidence Interval Coverage Probability}
\newacronym{QR}{QR}{Quantile Regression}
\newacronym{NSGA-II}{NSGA-II}{Non-dominated Sorting Genetic Algorithm II}
\newacronym{RBFNs}{RBFNs}{Radial Basis Function Networks}
\newacronym{XGBoost}{XGBoost}{eXtreme Gradient Boosting}
\newacronym{KDE}{KDE}{kernel density estimation}
\newacronym{GP-UCB}{GP-UCB}{Gaussian Process Upper Confidence Bound}
\newacronym{GPR}{GPR}{Gaussian Process Regression}

\newacronym{MSE}{MSE}{Mean Squared Error}
\newacronym{MAE}{MAE}{Mean Absolute Error}
\newacronym{PMV}{PMV}{Predicted Mean Vote}
\newacronym{MCD}{MCD}{Monte Carlo Dropout}
\newacronym{BNN}{BNN}{Bayesian Neural Network}

\title{Uncertainty-Aware Offline Data-Driven Multi-Objective Optimization}

\author{
Huanbo Lyu$^1$\and
Miqing Li$^1$\and
Shiqiao Zhou$^1$\and
Daniel Herring$^2$\and 
Jelena Ninic$^3$\and
Zheming Zuo$^1$\and
Lingfeng Wang$^4$\and
James Andrews$^4$\and
Fabian Spill$^4$\and
Shuo Wang$^1$\\
\affiliations
$^1$School of Computer Science, University of Birmingham, UK\\
$^2$Coventry University, UK\\
$^3$Department of Engineering, Durham University, UK\\
$^4$School of Mathematics, University of Birmingham, UK\\
\emails
hxl099@student.bham.ac.uk, m.li.8@bham.ac.uk, sxz363@student.bham.ac.uk, ae6648@coventry.ac.uk, jelena.ninic@durham.ac.uk, 
z.zuo.1@bham.ac.uk, lxw207@student.bham.ac.uk, j.w.andrews@bham.ac.uk, f.spill@bham.ac.uk, s.wang.2@bham.ac.uk
}

\begin{document}

\maketitle

\begin{abstract}
In offline data-driven multi-objective optimization (MOO), optimization is performed using surrogate models trained only on an offline dataset. These surrogate models contain inherent errors and uncertainty. This epistemic uncertainty can lead to incorrect dominance judgments, thereby misleading the search process. Existing methods mitigate this issue by incorporating uncertainty estimates from Gaussian Process Regression (GPR) to correct dominance judgments; however, they are restricted to GPR, and their optimization strategies cannot be scaled to other uncertainty quantification methods. In addition, GPR-based surrogates suffer from high computational cost.
We propose a simple yet effective dual-ranking strategy that flexibly leverages both predictive results and uncertainty estimates from different surrogate models. By performing non-dominated sorting on candidate solutions using both surrogate-based fitness values and uncertainty-aware fitness values, the proposed method prioritizes candidate solutions that are simultaneously high-quality and reliable. Through extensive experimental evaluations, including ablation, sensitivity, and comparative experiments, we demonstrate the effectiveness and robustness of the proposed dual-ranking strategy working with different surrogates. Our dual-ranking framework offers more robust solutions for data-limited, real-world applications. The code is available at https://anonymous.4open.science/r/dual-ranking-1D78/.
\end{abstract}

\section{Introduction}
In real-world applications, many \gls{MOPs}, namely, those involving multiple objectives to be optimized, lack analytical functions or simulation models that allow for direct evaluation~\cite{ddeo_rvea}. Consequently, data-driven \gls{MOEAs} have gained prominence by combining learning-based techniques with robust evolutionary search capabilities. These methods have proven effective in various applications, including trauma systems, airfoil design, energy systems, and smart building optimization \cite{ddeo_1,ddeo_2,ddeo_ibea,building_opt2}.

\begin{figure}[htbp]
    \centering
    \includegraphics[width=0.95\linewidth]
    {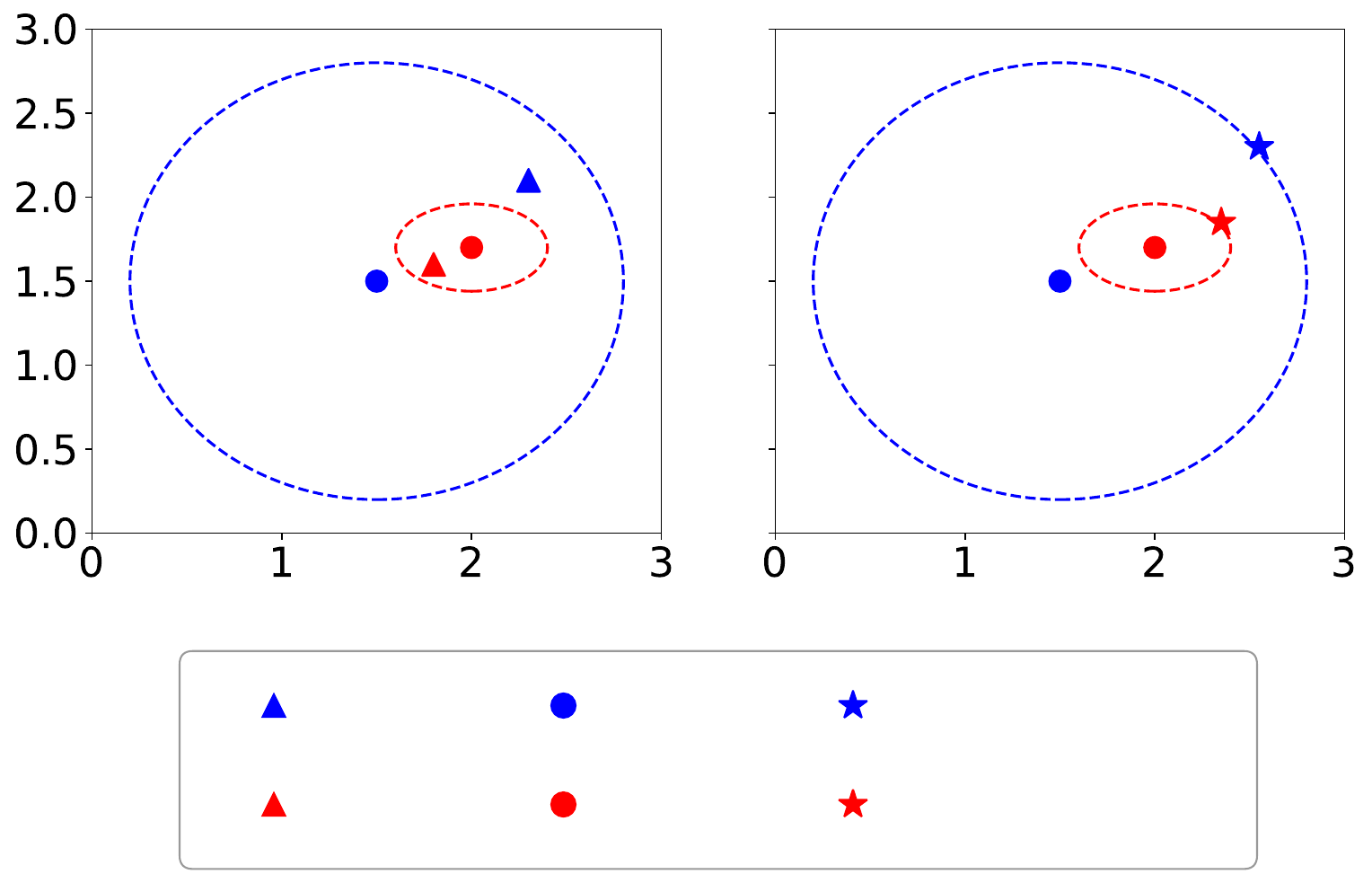}
    \put(-170,43){$f_1$}
    \put(-57,43){$f_1$}
    \put(-239,100){$f_2$}
    \put(-180,26){$f_\text{real}(\boldsymbol{A})$}
    \put(-180,10){$f_\text{real}(\boldsymbol{B})$}
    \put(-130,26){$f_\text{sur}(\boldsymbol{A})$}
    \put(-130,10){$f_\text{sur}(\boldsymbol{B})$}
    \put(-81,26){$f_\text{UA-sur}(\boldsymbol{A})$}
    \put(-81,10){$f_\text{UA-sur}(\boldsymbol{B})$}
\caption{An example of misleading dominance ranking in a minimization problem induced by surrogate-based evaluations and its correction via uncertainty-aware fitness values. The dashed circle denotes the predictive uncertainty interval centered at the surrogate-predicted objective values. Left: Relying solely on surrogate-based evaluations ($f_\text{sur}(\boldsymbol{A})$ and $f_\text{sur}(\boldsymbol{B})$) leads to an incorrect dominance judgment, $\boldsymbol{A} \prec \boldsymbol{B}$, which contradicts the dominance judgment under real evaluations, $\boldsymbol{B} \prec \boldsymbol{A}$. Right: Our method corrects this by incorporating uncertainty estimates, resulting in $\boldsymbol{B} \prec \boldsymbol{A}$, which will be explained in Section 3.}
\label{fig_1}
\end{figure}

Despite their success, offline data-driven \gls{MOEAs} face a critical challenge. Objective values are solely estimated via surrogates (\emph{i.e.}, data-driven models). Unlike surrogate-assisted optimization, no further real evaluations are available to improve the prediction accuracy. Even with a noise-free offline dataset where aleatoric uncertainty can be neglected, surrogate models still suffer from epistemic uncertainty arising from limited or biased training data, model selection, and training procedures. This approximation uncertainty can severely degrade optimization performance~\cite{ddeo_2} by misleading the search, resulting in suboptimal solutions \cite{optimizer_curse}. As shown in Figure~\ref{fig_1}, relying solely on surrogate-based evaluations $f_{\text{sur}}$ can lead to an incorrect dominance judgment, whereas incorporating uncertainty estimates can correct it in such cases.

To mitigate this issue, existing approaches include reducing approximation uncertainty through ensemble methods, and quantifying the uncertainty for incorporation into the optimization process. For the first approach, various ensemble methods have been proposed, such as selective surrogate ensembles~\cite{ddeo_2}, perturbation-based ensemble surrogates~\cite{ddeo_3} and ensemble surrogates with tri-training~\cite{ddeo_4}. However, these methods do not explicitly incorporate uncertainty information into the optimization process. Consequently, the accumulation of prediction errors during optimization can largely compromise the search process, potentially leading to suboptimal outcomes~\cite{ddeo_ibea}.

For the second approach, Kriging, also known as \gls{GPR}, is commonly adopted as the surrogate model due to its ability to provide uncertainty estimates~\cite{gpr}. Several methods leverage it, such as IBEA-MS~\cite{ddeo_ibea}, Prob-RVEA and Prob-MOEA/D~\cite{ddeo_rvea}. In addition, TGPR-MO~\cite{ddeo-TGPR} combines regression trees with \gls{GPR} to handle large datasets. DDMOEA/GAN~\cite{ddea_gan} utilizes discriminator and generator mechanisms of a generative adversarial network (GAN) for trust-aware evaluation and data augmentation, respectively.

In this vein, existing studies often rely on the predictive mean and standard deviation from \gls{GPR} models to adjust model selection or perform Monte Carlo sampling. However, these methods are typically restricted to \gls{GPR} and may suffer from high computational cost when the surrogate is trained on a large dataset. We consider a different perspective that directly incorporates uncertainty estimates into the fitness assignment procedure of a \gls{MOEA}. Our contributions are summarized as follows:

\begin{itemize}
  \item \textbf{Dual-Ranking Strategy.}
  To prioritize solutions that are both high-quality and reliable, while avoiding misleading rankings, we propose a dual-ranking strategy that considers the non-dominated front using both the surrogate-based function $f_{\text{sur}}$ and the uncertainty-aware function $f_{\text{UA-sur}}$. The proposed strategy can be flexibly integrated with different surrogate models.
  
  \item \textbf{Auto-$\alpha$ Selection Algorithm.}
  To align the uncertainty estimates from GPR with the quantile-based function, we design an auto-$\alpha$ selection algorithm to determine the hyperparameter $\alpha$ based on the target coverage $\tau$.
  
  \item \textbf{Extensive Evaluation.}
  We integrate the proposed dual-ranking strategy with different surrogate models and conduct comparative experiments against baseline methods. We also conduct ablation studies and sensitivity analyses of the proposed strategy, demonstrating its effectiveness and robustness under different settings.
\end{itemize}

\section{Preliminaries}
\subsection{Multi-Objective Optimization}
The \gls{MOPs} are considered in the following form:
\begin{align}\label{eq:mops}
\text{minimize }& \boldsymbol{f}(\boldsymbol{x}) = \left(f_1(\boldsymbol{x}), f_2(\boldsymbol{x}), \ldots, f_K(\boldsymbol{x})\right) \nonumber \\
\text{subject to }& \boldsymbol{x} \in S,
\end{align}
where $\boldsymbol{x}$ is the decision vector, $K \geq 2$ is the total number of objectives, and $S$ is the feasible region in the decision space. A solution $\boldsymbol{x}_{1} \in S$ dominates another solution $\boldsymbol{x}_{2} \in S$, denoted as $\boldsymbol{x}_{1} \prec \boldsymbol{x}_{2}$, if $f_k(\boldsymbol{x}_{1}) \leq f_k(\boldsymbol{x}_{2})$ for all $k = 1,\ldots,K$ and $f_k(\boldsymbol{x}_{1}) < f_k(\boldsymbol{x}_{2})$ for at least one $k = 1,\ldots,K$. The set of solutions that cannot be dominated by any other feasible solution is the \gls{PS}, and the set of their objective values is the \gls{PF}.

\subsection{Offline Data-Driven vs. Surrogate-Assisted Multi-Objective Optimization}
In offline data-driven \gls{MOO}, there is no explicit mathematical expression of the objective functions. Unlike surrogate-assisted optimization, which allows real evaluations to acquire new data points for improving the surrogate model during the optimization process \cite{kriging_assisted}, offline data-driven \gls{MOO} works in situations where querying solutions on the fly is not accessible \cite{ddeo_ibea,ddeo_rvea}, and thus deals with a pre-collected dataset. This dataset remains unchanged during optimization. The objective functions are approximated by training surrogates on this offline dataset. 

\subsection{Surrogate Modeling}
The surrogate $\mathcal{M}$, trained on the offline dataset $\mathcal{D}$, aims to predict the objective values of a candidate solution $\boldsymbol{x}$. The $k$-th objective value $f_{k}(\boldsymbol{x})$ estimated from the surrogate $\mathcal{M}_{k}$ can be expressed as:
\begin{align}\label{eq:surrogate_1}
f_{k}(\boldsymbol{x}) \approx \mathcal{M}_{k} (\boldsymbol{x}).
\end{align}

Existing mainstream surrogate models that can provide both predictions and uncertainty-related information include \gls{GPR} \cite{gpr}, \gls{QR} \cite{quantile_regression}, \gls{BNN} \cite{bnn}, and ensemble methods \cite{ddeo_2}. \gls{GPR} is widely used in surrogate-assisted and offline data-driven \gls{MOO} \cite{jin_gpr}, as it naturally provides predictive mean and variance. However, standard GPR suffers from high computational cost and degraded performance in high dimensions. QR does not rely on Gaussian assumptions and is useful under asymmetric or heavy-tailed distributions. BNN utilizes the strong representation ability of neural networks and is scalable to larger datasets and higher-dimensional problems, although training and inference are computationally expensive. Ensemble methods are widely adopted in engineering problems, and their uncertainty is commonly derived from multiple independently trained surrogates rather than the posterior distribution. In this work, we consider GPR, QR, and BNN as surrogate models.

\begin{figure*}[!htbp]
\includegraphics[width=\textwidth]{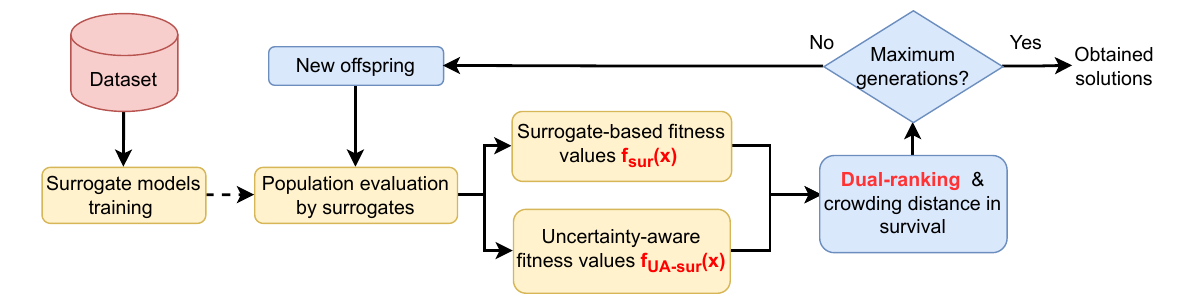}
\caption{Workflow of surrogate training and optimization. Dual-ranking is implemented by performing non-dominated sorting on solutions using both the surrogate-based function $f_{\text{sur}}$ and the uncertainty-aware function $f_{\text{UA-sur}}$.}
\label{fig: flow_chart}
\end{figure*}

\subsubsection{Gaussian Process Regression}
The \gls{GPR} model is fully specified by a mean function $\mu(\mathbf{x})$ and a covariance (kernel) function $k(\mathbf{x}, \mathbf{x}')$:
\begin{equation}
f(\mathbf{x}) \sim \mathcal{GP}(\mu(\mathbf{x}), k(\mathbf{x}, \mathbf{x}')).
\end{equation}
In this work, we adopt the Gaussian kernel \cite{gaussian_kernel} and the  Automatic Relevance Determination (ARD) Mat\'ern 5/2 kernel \cite{kernel_Matern_52}, as the latter does not make the covariance functions unrealistically smooth \cite{ddeo-TGPR}:
\begin{align}
k(\mathbf{x}, \mathbf{x}')
=&
\sigma_f^2
\exp\!\left(-\tfrac{1}{2} r^2(\mathbf{x}, \mathbf{x}')\right), \\ \nonumber
k_{\mathrm{M52}}(\mathbf{x}, \mathbf{x}')
=&
\sigma_f^2
\left(1 + \sqrt{5r^2(\mathbf{x}, \mathbf{x}')}
+ \tfrac{5}{3} r^2(\mathbf{x}, \mathbf{x}')\right) \nonumber \\ 
&\exp\!\left(-\sqrt{5r^2(\mathbf{x}, \mathbf{x}')}\right), 
\end{align}
where $r^2(\mathbf{x}, \mathbf{x}')
= \sum_{d=1}^{D} \frac{(x_d - x'_d)^2}{\ell_d^2}$ and $x_d$ and $x'_d$ are the $d$-th components of the decision vector of $\mathbf{x}$ and $\mathbf{x}'$. The parameters $\sigma_f^2$ and $\ell_d$ represent the amplitude, and length scale of the $d$ th variable.

\subsubsection{Quantile Regression}
\gls{QR} estimates the conditional quantile (\emph{e.g.} median) of a target variable. For a given quantile level $\tau \in (0, 1)$, the estimator is obtained by:
\begin{align}
\hat{\boldsymbol{\beta}}_\tau = \arg\min_{\boldsymbol{\beta}} \sum_{i=1}^n \rho_\tau \left( y_i - \boldsymbol{x}_i^\top \boldsymbol{\beta} \right),
\end{align}
where $\boldsymbol{\beta}$ represents the regression coefficients and the loss function $\rho_\tau(y,\hat{y})$ is the ``check function'' \cite{pinball_loss}, defined as
\begin{align}
\rho_\tau(y,\hat{y}) =
\begin{cases}
\tau \cdot (y-\hat{y}) & \text{if } \hat{y} \leq y \\
(1 - \tau) \cdot (\hat{y}-y) & \text{if } \hat{y} > y.
\end{cases}
\end{align}
The simplicity and generality of this formulation makes quantile regression widely applicable across various domains \cite{conformal_quantile_regression}. We employ \gls{QR} to estimate uncertainty intervals by predicting multiple
conditional quantiles.

\subsubsection{Bayesian Neural Network}
\gls{BNN} places probability distributions over the network weights instead of point estimates, enabling the modeling of epistemic uncertainty. \gls{BNN} aims to approximate the posterior distribution $p(\mathbf{w} \mid \mathcal{D})$ over weights $\mathbf{w}$ using variational inference. For a given input $\boldsymbol{x}$, the predictive distribution is approximated as:
\begin{align}
p(y \mid \boldsymbol{x},\mathcal{D}) \approx \frac{1}{S} \sum_{s=1}^{S} p(y \mid \boldsymbol{x}, \mathbf{w}^{(s)}) \quad \mathbf{w}^{(s)} \sim q(\mathbf{w}),
\end{align}
where $q(\mathbf{w})$ is the learned variational posterior.

\section{Method}
In this section, we propose the dual-ranking strategy that leverages the results from both the surrogate-based function $f_{\text{sur}}$ and the uncertainty-aware function $f_{\text{UA-sur}}$ in the fitness assignment procedure of a \gls{MOEA}, namely \gls{NSGA-II} \cite{nsga_ii}, which is one of the most widely used algorithms in \gls{MOO}.

\subsection{Motivation}
We assume that data points $x_1, x_2, \ldots, x_n$ in a dataset $\mathcal{D}$ are drawn from an unknown distribution $p(x)$. At the model training stage, the surrogates aim to reduce the approximation errors over the dataset. However, at the optimization stage, an off-the-shelf multi-objective optimization algorithm (e.g., NSGA-II) explores the whole input space $\boldsymbol{X}$ to minimize the objective $\boldsymbol{f}(\boldsymbol{x})$, relying solely on the surrogate-based evaluation. When the optimizer explores unseen areas (\emph{i.e.} the candidate solution $x$ lies in a low-density region of the distribution $p(x)$), the predicted objective values may be unrealistically small with high uncertainty. Such solutions will be falsely preferred by the algorithm as they are regarded as non-dominated solutions, as illustrated in Figure~\ref{fig_1}. To address this issue, we consider both predicted values and uncertainty estimates from surrogates. The detailed workflow is shown in Figure \ref{fig: flow_chart}.

\subsection{Uncertainty-Aware Function}
For the $k$-th objective of a candidate solution $\boldsymbol{x}$, the surrogate model $\mathcal{M}_k$ outputs both a predicted value and an uncertainty estimate, whose specific forms depend on the regression model:
\begin{align}
&\mathcal{M}_{\text{GPR},k}(\boldsymbol{x})
= \big( \mu_k(\boldsymbol{x}),\, \sigma_k^2(\boldsymbol{x}) \big), \\
&\mathcal{M}_{\text{QR},k}(\boldsymbol{x})
= \big( q_{\text{med},k}(\boldsymbol{x}),\, q_{\text{up},k}(\boldsymbol{x}) \big), \\
&\mathcal{M}_{\text{BNN},k}(\boldsymbol{x})
= \big(
q_{\text{med},k}(\boldsymbol{x}),
q_{\text{up},k}(\boldsymbol{x})
\big),
\end{align}
where the predictive mean $\mu_k(\boldsymbol{x})$ and the median prediction $q_{\text{med},k}(\boldsymbol{x})$ ($\tau=0.5$) denote the predicted value, predictive variance $\sigma_k^2(\boldsymbol{x})$ and the interval induced by the quantile estimates $q_{\text{up},k}(\boldsymbol{x})$ indicate the predictive uncertainty. For BNN, the predictive quantiles are computed from Monte Carlo samples drawn from the variational posterior over network weights.

During the fitness assignment procedure of the optimization, the fitness value of a candidate solution $\boldsymbol{x}$ is derived from the surrogate-based function $f_{\text{sur},k}$, where the results are the predictions from the surrogate model:
\begin{align}
f_{\text{sur},k}(\boldsymbol{x}) =
\begin{cases}
\mu_k(\boldsymbol{x}) & \text{(GPR)} \\
q_{\text{med},k}(\boldsymbol{x}) & \text{(QR)} \\
q_{\text{med},k}(\boldsymbol{x}) & \text{(BNN)}.
\end{cases}
\end{align}

To reduce the risk of misleading dominance judgments, we draw inspiration from the \gls{GP-UCB} principle~\cite{opt_ucb} and define an uncertainty-aware function $f_{\text{UA-sur},k}$. The corresponding uncertainty-aware fitness values are defined as follows:
\begin{align}
f_{\text{UA-sur},k}(\boldsymbol{x}) =
\begin{cases}
\mu_k(\boldsymbol{x}) + \alpha\, \sigma_k(\boldsymbol{x}) & \text{(GPR)} \\
q_{\text{up},k}(\boldsymbol{x}) & \text{(QR)} \\
q_{\text{up},k}(\boldsymbol{x}) & \text{(BNN)},
\end{cases}
\end{align}
where $\alpha$ is a hyperparameter that controls the degree to which solutions are penalized, while the upper quantile $q_{\text{up},k}$ characterizes the predictive uncertainty and is determined by the selected quantile level $\tau$ (\emph{e.g.} $\tau=0.9$). Larger uncertainty estimates are typically associated with less reliable fitness evaluations, which increases the risk of incorrectly judging dominance among solutions. Therefore, in minimization problems, the uncertainty-aware function penalizes large uncertainty estimates.

Inspired by the \gls{CICP} indicator \cite{cicp}, we design an auto-$\alpha$ selection algorithm to select the hyperparameter $\alpha$, as shown in Algorithm~\ref{alg:find_alpha}. This hyperparameter is introduced to align the degree of penalization between uncertainty estimates provided by GPR-based models and the uncertainty-related predictions obtained from quantile-based models. The algorithm iteratively adjusts $\alpha$ on a pre-collected validation set until the empirical coverage $C(\alpha)$ of the upper confidence bound reaches the given target coverage $\tau$. Here, the target coverage $\tau$ is set equal to the selected quantile level $\tau$ used in $q_{\text{up},k}$.

\begin{algorithm}[htbp]
\caption{Auto-$\alpha$ Selection}
\label{alg:find_alpha}
\begin{algorithmic}[1]
\STATE \textbf{Input}: Validation set $(\mathbf{X}, \mathbf{y})$ with $N_{\text{val}}$ samples, surrogate model $\mathcal{M}$, target coverage $\tau$
\STATE \textbf{Output}: $\alpha$

\STATE $(\boldsymbol{\mu}, \boldsymbol{\sigma}) \leftarrow \mathcal{M}(\mathbf{X})$
\FOR{$\alpha = 0$ \TO $\alpha_{\max}$}
    \STATE $f_{\text{UA-sur}} \leftarrow \boldsymbol{\mu} + \alpha \boldsymbol{\sigma}$
    \STATE $C(\alpha) = \frac{1}{N_{\text{val}}} \sum_{i=1}^{N_{\text{val}}} \mathbb{I}\!\left(y_i \le f_{\text{UA-sur},i}\right)$
    \IF{$C(\alpha) \ge \tau$}
        \RETURN $\alpha$
    \ENDIF
\ENDFOR
\end{algorithmic}
\end{algorithm}

\begin{proposition}[Monotonicity and existence of $\alpha$]
\label{prop:alpha}
On the validation set, the empirical coverage $C(\alpha)$ is a non-decreasing function of $\alpha$.
\[
C(\alpha)
= \frac{1}{N_{\mathrm{val}}}
\sum_{i=1}^{N_{\mathrm{val}}}
\mathbb{I}\!\left(
y_i \le \mu_i + \alpha \sigma_i
\right)
\]

Moreover, if there exists $\alpha^\star$ such that
$C(\alpha^\star) \ge \tau$,
Algorithm~\ref{alg:find_alpha} is guaranteed to return a feasible
$\alpha \le \alpha^\star$.
\end{proposition}

\subsection{Dual-Ranking Strategy}
In the survival step, for the $k$-th objective, each candidate solution is evaluated using both the surrogate-based function $f_{\text{sur},k}$ and the uncertainty-aware function $f_{\text{UA-sur},k}$. The dual-ranking strategy aims to select solutions with both good objective values and low uncertainty estimates. As shown in Algorithm \ref{alg:dual_ranking}, for each candidate solution $\boldsymbol{x}$ in population $P$, the surrogate-based fitness values $f_{\text{sur}}(\boldsymbol{x}) \in \mathbb{R}^{M}$ and the uncertainty-aware fitness values $f_{\text{UA-sur}}(\boldsymbol{x}) \in \mathbb{R}^{M}$ are concatenated to form the hybrid fitness values $f_{\text{hyb}}(\boldsymbol{x}) \in \mathbb{R}^{2M}$. Non-dominated sorting is then applied to the population $P$ with updated fitness values $f_{\text{hyb}}(\boldsymbol{x}) \in \mathbb{R}^{2M}$ of each candidate solution to obtain the non-dominated fronts. 

\begin{algorithm}[htbp]
\caption{Dual-Ranking}
\label{alg:dual_ranking}
\begin{algorithmic}[1]
\STATE \textbf{Input}: Population $P$, fitness value $f_{\text{sur}}(\boldsymbol{x}) \in \mathbb{R}^{M}$ and uncertainty-aware fitness value $f_{\text{UA-sur}}(\boldsymbol{x}) \in \mathbb{R}^{M}$ of a candidate solution $\boldsymbol{x}$
\STATE \textbf{Output}: Non-dominated front $\mathcal{F}$
\FOR{each $\boldsymbol{x}$ in $P$}
  \STATE $f_{\text{hyb}}(\boldsymbol{x}) \in \mathbb{R}^{2M} 
\gets \textit{Concatenate}(f_{\text{sur}}(\boldsymbol{x}), f_{\text{UA-sur}}(\boldsymbol{x}))$
\ENDFOR
\STATE $\mathcal{F} \gets \textit{NonDominatedSorting}(P)$
\end{algorithmic}
\end{algorithm}

\begin{proposition}[Uncertainty-aware dominance judgment] \label{proposition_2}
Consider two candidate solutions $x_a$ and $x_b$ in a minimization problem.
Assume that $x_a \prec x_b$ in the surrogate-based objective space
$f_{\text{sur}}$, i.e.,
\[
\forall k, f_{\text{sur},k}(x_a) \le f_{\text{sur},k}(x_b), 
\]
\[
\exists k_j,
f_{\text{sur},k_j}(x_a) < f_{\text{sur},k_j}(x_b).
\]
If there exists at least one objective $k$ such that
\[
f_{\text{UA-sur},k}(x_a) > f_{\text{UA-sur},k}(x_b),
\]
then $x_a \not\prec x_b$ in the hybrid objective space
$f_{\text{hyb}} \in \mathbb{R}^{2M}$.
\end{proposition}

\subsubsection{Time Complexity}
We analyze the time complexity of the proposed dual-ranking strategy when used with \gls{NSGA-II}. The standard non-dominated sorting procedure in \gls{NSGA-II} has a time complexity of $O(MN^2)$, where $N$ denotes the number of candidate solutions and $M$ the number of objectives. In the proposed dual-ranking strategy, the time complexity of concatenation operator is $O(MN)$. Non-dominated sorting is performed in the $2M$-dimensional objective space, resulting in a worst-case time complexity of $O(2MN^2)$. Therefore, total time complexity is $O(MN) + O(2MN^2) \approx O(MN^2)$.

The benefits of the proposed dual-ranking strategy are summarized as follows. Our method jointly considers both predictive values and uncertainty-aware predictive values from surrogates to mitigate misleading rankings, while prioritizing candidate solutions that are both high-quality and reliable. Solutions with good predicted values but high uncertainty estimates or uncertainty-related predictions are penalized during ranking, preventing them from being assigned overly good ranks based solely on optimistic fitness values. Moreover, the proposed strategy can be easily integrated into non-dominated sorting, has the potential to work with other fitness assignment strategies. It can also be flexibly integrated with different surrogate models, making it suitable for offline datasets with different characteristics.

\section{Experiments}
\subsection{Experimental Setup}
\subsubsection{Benchmark Problems}
We consider 11 benchmark problems for the comparative experiments, including 8 unconstrained and 3 constrained problems. The unconstrained benchmarks consist of DTLZ1–DTLZ7~\cite{opt_dtlz} with 2 objectives and 10 decision variables, and Omnitest~\cite{omnitest} with 2 objectives and 2 decision variables. The constrained benchmarks include BNH~\cite{bnh}, Truss2D~\cite{truss2d}, and Welded Beam~\cite{welded_beam}.

\subsubsection{Baseline Methods} \label{sec:baseline}
The baseline methods include Prob-RVEA, Prob-MOEA/D \cite{ddeo_rvea}, IBEA-MS \cite{ddeo_ibea}, TGPR-MO \cite{ddeo-TGPR}, and DDMOEA/GAN \cite{ddea_gan}. We adopt the default or recommended hyperparameter settings from the open source code. The experiments are performed over 30 independent runs, with fixed random seeds ranging from 1 to 30 unless otherwise specified.

Different baseline methods adopt different optimization algorithms combined with the GPR surrogate model. As some types of optimization algorithms may be better suited for handling specific benchmark problems, the comparison experiments presented in Section~\ref{sec:results_baseline} mainly focus on overall optimization performance across all benchmarks.

\subsubsection{Offline Dataset} 
The offline data are generated using Latin Hypercube Sampling (LHS)~\cite{sampling}, and the random seed is fixed to 42 to ensure identical data samples. Following the default setting in~\cite{ddeo_ibea}, the number of samples for limited dataset is set to $11D - 1$, where $D$ denotes the number of decision variables. We also test the methods on a relatively large dataset with 1,000 samples, where the random seed is fixed to 42, and report the detailed results in the Supplementary Material.

\subsubsection{Surrogate Models}
For \gls{GPR}, we use the Gaussian kernel \cite{gaussian_kernel} and ARD Mat\'ern 5/2 kernel \cite{kernel_Matern_52}. In some test cases, to avoid constant predictions, the variance of the Gaussian noise is set to a very small value of $10^{-6}$. \gls{QR} is implemented using AutoGluon-Tabular \cite{ml_autogluon}, with quantile levels of 0.5 and 0.9, and a fixed random seed of 42. \gls{BNN} is implemented by Pyro~\cite{pyro} and employs the AutoDiagonalNormal for variational inference. The BNN model uses a hidden dimension of 16, an initial learning rate of $10^{-3}$, the Adam optimizer~\cite{adam}. The number of training epochs is tuned with early stopping mechanism for each benchmark problem. Other settings are reported in the Supplementary Material.

\subsubsection{Optimization Algorithms}
The optimization framework is implemented using pymoo \cite{pymoo}. The population size is set to 100, and the maximum number of generations is 100, resulting in a total of 10,000 fitness evaluations. For the reproduction operators, Simulated Binary Crossover (SBX) \cite{opt-sbx} is applied with a crossover probability of 1.0 and a distribution factor of $\eta = 20$. Polynomial mutation \cite{nsga_ii} is employed with a mutation probability of $1/D$ and the same distribution factor. 

\subsubsection{Performance Indicators}
For prediction performance of surrogates, the indicators include \gls{MSE} and \gls{CICP} \cite{cicp}, and the detailed results are provided in the Supplementary Material. \gls{MSE} is used to compare the difference between the prediction $\hat{y}$ and the ground truth $y$. \gls{CICP} evaluates uncertainty estimates by computing the fraction of function values that fall inside the corresponding confidence intervals. 

For optimization, the indicators include \gls{MSE}, \gls{IGD+} \cite{igd_plus} and \gls{HV} \cite{hypervolume}. The \gls{MSE} indicator measures the difference between the real evaluations and the surrogate-based evaluations. Both \gls{IGD+} and \gls{HV} indicators are employed to evaluate the optimization results of real evaluations, since the surrogate-based evaluations may introduce errors. The \gls{IGD+} indicator is available as the \gls{PF} is known on synthetic problems. For the \gls{HV} indicator, Min-Max normalization \cite{indicator} is applied, and the bounds are selected based on the benchmark \cite{benchmark} and valid ranges, with details provided in the Supplementary Material. The reference point is $(1.1, 1.1)$. 

\subsection{Surrogate Prediction Performance Results}
The surrogate prediction performance results are summarized in this section, while detailed results and discussions for both the limited dataset and the 1,000-sample dataset are provided in the Supplementary Material. The GPR (Mat\'ern) model generally achieves lower prediction errors than the GPR (RBF) model across most benchmark problems. The GPR (RBF) model occasionally yields slightly higher \gls{CICP} values but increased \gls{MSE}. Autogluon-QR consistently delivers high \gls{CICP} values across nearly all cases, demonstrating robust uncertainty estimates despite relatively larger MSE. The BNN model shows larger variability in both prediction accuracy and uncertainty estimates. 

These observations indicate a trade-off between prediction accuracy and reliable uncertainty estimates, which motivates the uncertainty-aware function with dual-ranking strategy adopted in this work.

\subsection{Ablation Study and Sensitivity Analysis}
To demonstrate the effectiveness of the proposed dual-ranking strategy, we conduct an ablation study and a sensitivity analysis across 11 benchmark problems. These analyses compare the optimization performance between baseline methods and their dual-ranking variants. For the sensitivity analysis, the target coverage rate is set to $\tau = 0.80, 0.90, 0.95$ for GPR-based models, while the quantile level is set to $\tau = 0.80, 0.90, 0.95$ for AutoGluon-QR and BNN models. Due to page limits, a summary of the comparison counts is shown in Table~\ref{tab:opt_counts}, and the detailed results are reported in the Supplementary Material.

\begin{table}[htbp]
\centering
\setlength{\tabcolsep}{3pt}
\begin{tabular}{l|ccc|ccc|ccc}
\hline
\multirow{2}{*}{Methods} 
& \multicolumn{3}{c|}{MSE$\downarrow$} 
& \multicolumn{3}{c|}{IGD+$\downarrow$} 
& \multicolumn{3}{c}{HV$\uparrow$} \\
 & $+$ & $=$ & $-$ & $+$ & $=$ & $-$ & $+$ & $=$ & $-$ \\
\hline
GPR (RBF) + DR ($\tau{=}0.80$) & 8 & 0 & 3 & 6 & 0 & 5 & 3 & 6 & 2 \\
GPR (RBF) + DR ($\tau{=}0.90$) & 9 & 0 & 2 & 5 & 0 & 6 & 4 & 4 & 3 \\
GPR (RBF) + DR ($\tau{=}0.95$) & 7 & 0 & 4 & 8 & 0 & 3 & 6 & 2 & 3 \\
\hline
GPR (Mat\'ern) + DR ($\tau{=}0.80$) & 6 & 0 & 5 & 5 & 0 & 6 & 5 & 3 & 3 \\
GPR (Mat\'ern) + DR ($\tau{=}0.90$) & 6 & 0 & 5 & 6 & 0 & 5 & 5 & 3 & 3 \\
GPR (Mat\'ern) + DR ($\tau{=}0.95$) & 6 & 0 & 5 & 5 & 0 & 6 & 5 & 3 & 3 \\
\hline
QR + DR ($\tau{=}0.80$) & 6 & 0 & 5 & 7 & 0 & 4 & 7 & 2 & 2 \\
QR + DR ($\tau{=}0.90$) & 7 & 0 & 4 & 6 & 0 & 5 & 5 & 4 & 2 \\
QR + DR ($\tau{=}0.95$) & 7 & 0 & 4 & 6 & 0 & 5 & 6 & 4 & 1 \\
\hline
BNN + DR ($\tau{=}0.80$) & 6 & 1 & 4 & 4 & 3 & 4 & 4 & 5 & 2 \\
BNN + DR ($\tau{=}0.90$) & 6 & 1 & 4 & 4 & 4 & 3 & 4 & 5 & 2 \\
BNN + DR ($\tau{=}0.95$) & 7 & 1 & 3 & 4 & 3 & 4 & 4 & 4 & 3 \\
\hline
\end{tabular}
\caption{Ablation study and sensitivity analysis results across 11 benchmark problems, for different surrogate models combined with the NSGA-II optimization algorithm (omitted). Each method is evaluated under different Dual-Ranking (DR) settings for sensitivity analysis. 
All indicators are optimization performance indicators. Lower is better for \gls{MSE} and \gls{IGD+}, while higher is better for \gls{HV}. For each indicator, ``$+$/$=$/$-$'' indicates better/comparable/worse performance than the corresponding baseline without the dual-ranking strategy.}
\label{tab:opt_counts}
\end{table}

\begin{table*}[!htbp]
\centering
\small
\setlength{\tabcolsep}{3pt} 
\begin{tabular}{l ccc ccc ccc ccc}
\toprule
\multirow{2}{*}{Methods} 
& \multicolumn{3}{c}{DTLZ1} & \multicolumn{3}{c}{DTLZ2} & \multicolumn{3}{c}{DTLZ3} & \multicolumn{3}{c}{DTLZ4} \\
\cmidrule(lr){2-4} \cmidrule(lr){5-7} \cmidrule(lr){8-10} \cmidrule(lr){11-13}
& MSE & IGD+ & HV & MSE & IGD+ & HV & MSE & IGD+ & HV & MSE & IGD+ & HV \\
\midrule
Prob-RVEA    & 1.09E+05 & 2.70E+02 & 0.67 & 2.24E-02 & 7.78E-02 & 1.09 & 4.24E+05 & 5.54E+02 & 0.82 & 3.80E-02 & 3.25E-01 & 0.91 \\
Prob-MOEA/D  & 1.12E+05 & 2.95E+02 & 0.62 & 1.88E-02 & 1.12E-01 & 1.09 & 4.51E+05 & 5.94E+02 & 0.79 & 5.87E-03 & 4.75E-01 & 0.79 \\
IBEA-MS      & 3.99E+03 & 1.87E+02 & 1.06 & 7.38E-02 & 2.86E-02 & 1.11 & 1.10E+05 & 2.24E+02 & 1.19 & 1.45E-02 & 3.37E-01 & 0.86 \\
TGPR-MO      & 4.73E+03 & 1.63E+02 & 1.07 & 3.39E-02 & 2.98E-02 & 1.11 & 2.32E+04 & 4.82E+02 & 1.08 & 2.96E-01 & 3.96E-01 & 0.82 \\
DDMOEA/GAN   & 3.25E+03 & 8.45E+01 & 1.18 & 3.26E+00 & 1.72E+00 & 0.34 & 1.24E+04 & 2.29E+02 & 1.19 & 1.13E+00 & 9.33E-01 & 0.61 \\
GPR(RBF)+DR  & 9.43E+03 & 1.16E+02 & 1.14 & 5.44E-02 & 5.17E-01 & 0.97 & 2.28E+04 & 4.18E+02 & 1.11 & 2.08E-01 & 4.73E-01 & 0.83 \\
GPR(Mat\'ern)+DR  & 7.94E+03 & 1.33E+02 & 1.12 & 2.00E-02 & 6.61E-02 & 1.10 & 2.27E+04 & 4.24E+02 & 1.11 & 2.43E-01 & 4.43E-01 & 0.80 \\
QR+DR        & 6.63E+03 & 2.06E+02 & 1.01 & 5.23E-02 & 1.16E-01 & 1.08 & 3.85E+04 & 4.89E+02 & 1.10 & 6.00E-01 & 1.96E-01 & 1.01 \\
BNN+DR       & 8.72E+03 & 2.25E+02 & 0.98 & 5.72E-01 & 1.10E+00 & 0.72 & 3.70E+04 & 5.26E+02 & 1.08 & 6.67E-01 & 1.39E+00 & 0.50 \\
\midrule

Methods & \multicolumn{3}{c}{DTLZ5} & \multicolumn{3}{c}{DTLZ6} & \multicolumn{3}{c}{DTLZ7} & \multicolumn{3}{c}{Omnitest} \\
\cmidrule(lr){2-4} \cmidrule(lr){5-7} \cmidrule(lr){8-10} \cmidrule(lr){11-13}
& MSE & IGD+ & HV & MSE & IGD+ & HV & MSE & IGD+ & HV & MSE & IGD+ & HV \\
\midrule
Prob-RVEA    & 2.26E-02 & 8.13E-02 & 1.08 & 1.44E-02 & 7.77E+00 & 0.55 & 1.44E+00 & 9.07E-01 & 1.03 & 1.96E-01 & 1.80E-01 & 1.07 \\
Prob-MOEA/D  & 1.88E-02 & 1.24E-01 & 1.06 & 1.08E-02 & 8.06E+00 & 0.51 & 7.87E-01 & 9.12E-01 & 1.07 & 4.45E-02 & 5.70E-01 & 0.91 \\
IBEA-MS      & 7.38E-02 & 2.98E-02 & 1.09 & 5.17E+00 & 1.78E+00 & 1.14 & 5.70E-02 & 1.91E+00 & 1.11 & - & - & - \\
TGPR-MO      & 3.43E-02 & 2.91E-02 & 1.09 & 2.03E+00 & 3.73E+00 & 0.99 & 4.17E-07 & 2.55E-02 & 1.11 & 2.97E-01 & 2.22E-01 & 1.03 \\
DDMOEA/GAN   & 3.24E+00 & 2.02E+00 & 0.09 & 7.28E+00 & 5.57E+00 & 0.77 & 1.45E+01 & 2.78E-01 & 1.09 & 1.31E+00 & 2.48E-01 & 1.08 \\
GPR(RBF)+DR  & 5.44E-02 & 5.12E-01 & 0.93 & 6.32E-01 & 4.98E+00 & 0.89 & 1.13E-01 & 1.30E-01 & 1.10 & 4.37E-01 & 2.15E-01 & 1.06 \\
GPR(Mat\'ern)+DR  & 2.00E-02 & 6.68E-02 & 1.08 & 1.45E+00 & 4.51E+00 & 0.94 & 1.41E-07 & 4.21E-03 & 1.11 & 5.56E-01 & 1.95E-01 & 1.08 \\
QR+DR        & 5.23E-02 & 1.18E-01 & 1.06 & 2.15E-01 & 6.66E+00 & 0.68 & 3.56E+00 & 8.00E-02 & 1.10 & 3.70E-01 & 5.00E-01 & 0.91 \\
BNN+DR       & 4.97E-01 & 1.15E+00 & 0.63 & 9.03E-01 & 5.48E+00 & 0.83 & 1.85E+00 & 1.66E-01 & 1.10 & 1.78E+00 & 1.47E-01 & 1.12 \\
\bottomrule
\end{tabular}
\caption{Optimization performance comparison with baseline offline data-driven \gls{MOO} methods on 8 unconstrained benchmarks. Our method is combined with \gls{NSGA-II} (omitted) using the dual-ranking (DR) strategy with $\tau=0.9$. Lower is better for \gls{MSE} and \gls{IGD+}, while higher is better for \gls{HV}. The symbol ``-'' indicates that the experiment cannot be conducted under this setting.}
\label{tab:opt_baseline}
\end{table*}

For GPR (RBF), the dual-ranking strategy improves the reliability of surrogate-based evaluations, as evidenced by lower MSE values on 7 to 9 problems depending on the target coverage rate $\tau$. These gains partially translate into improved optimization performance (IGD+), while HV remains unchanged in many cases. For GPR (Mat\'ern), the dual-ranking strategy yields relatively stable performance across MSE, IGD+, and HV regardless of $\tau$. The AutoGluon-QR method achieves lower MSE and IGD+ values on 6–7 benchmark problems, along with consistent improvements in HV. For BNN, the dual-ranking strategy yields moderate improvements in MSE, but smaller improvements in IGD+ and HV.

Overall, the results indicate that the dual-ranking strategy is robust to the choice of $\tau$ within a reasonable range. The choice of $\tau$ mainly affects optimization performance across different problems, rather than altering the reliability of surrogate-based evaluations. It also suggests that a moderate value (e.g., $\tau=0.90$) provides a good trade-off and is used as the default setting in the following experiments.

We also conduct an ablation study comparing the dual-ranking strategy with the sole adoption of $f_\text{UA-sur}$, while the detailed results are reported in the Supplementary Material. We find that the dual-ranking strategy achieves more improvements (or ties) in IGD+ and HV than in MSE.

\subsection{Comparison Experiments with Baseline Methods} \label{sec:results_baseline}
This experiment compares the optimization performance of the proposed dual-ranking–based methods with baseline offline data-driven \gls{MOO} methods on 8 unconstrained benchmark problems, as most baseline methods cannot handle constraints. Table \ref{tab:opt_baseline} reports the detailed comparison results measured by the optimization performance indicators MSE, IGD+ and HV. Table~\ref{tab:opt_baseline_ranks} summarizes the overall ranking results, while detailed ranks are provided in the Supplementary Material. The ranks are computed for each performance indicator on each benchmark problem. The mean and standard deviation of the overall ranks are then calculated based on these detailed ranks.

\begin{table}[!h]
\centering
\begin{tabular}{lc}
\toprule
Methods & Overall ranks $\downarrow$\\
\midrule
Prob-RVEA          
& 5.17 $\pm$ 2.56 \\
Prob-MOEA/D        
& 6.12 $\pm$ 3.02 \\
\underline{IBEA-MS}            
& \underline{3.38 $\pm$ 2.66} \\
TGPR-MO            
& 3.50 $\pm$ 1.78 \\
DDMOEA/GAN         
& 6.00 $\pm$ 3.21 \\
GPR(RBF)+DR        
& 4.67 $\pm$ 1.57 \\
\textbf{GPR(Mat\'ern)+DR} 
& \textbf{3.25 $\pm$ 1.51} \\
QR+DR              
& 5.17 $\pm$ 1.80 \\
BNN+DR             
& 6.38 $\pm$ 2.14 \\
\bottomrule
\end{tabular}
\caption{Overall ranking results (mean $\pm$ std.) of different methods across 8 unconstrained benchmark problems. Our method is combined with the \gls{NSGA-II} (omitted) and the dual-ranking (DR) strategy with $\tau=0.9$. \textbf{Bold} indicates the best result, and \underline{underline} indicates the second-best result.}
\label{tab:opt_baseline_ranks}
\end{table}

As stated in Section \ref{sec:baseline} Baseline Methods, we focus on the overall optimization performance. Among all compared methods, GPR(Mat\'ern)+DR achieves the best overall rank (3.25 ± 1.51), followed by IBEA-MS (3.38 ± 2.66) and TGPR-MO (3.50 ± 1.78). These results indicate that incorporating the dual-ranking strategy into GPR with a Mat\'ern kernel leads to superior and more consistent optimization performance.

The standard deviation of overall ranks further highlights the robustness advantage of the proposed dual-ranking strategy. Although dual-ranking–based methods do not achieve the best performance on each benchmark problem, all surrogates incorporating the dual-ranking strategy exhibit relatively small standard deviations. In contrast, while IBEA-MS or DDMOEA/GAN may perform competitively on certain problems, their larger standard deviations indicate weaker robustness across different benchmarks. Overall, the proposed dual-ranking strategy demonstrates improved robustness across all combined surrogate models.

To further explore the optimization performance of the proposed methods, we conduct comparison experiments with baselines, where the surrogate models are trained on a relatively large dataset with 1,000 samples. Due to page limits, Table~\ref{tab:opt_baseline_1000_ranks} summarizes the overall ranking results, while detailed optimization performance and ranking results are provided in the Supplementary Material.

Compared with the ranking results in Table~\ref{tab:opt_baseline_ranks}, GPR(Mat\'ern)+DR achieves the best overall performance in both settings, demonstrating the robustness of the proposed strategy across different dataset sizes. Dual-ranking–based methods benefit from increased training data, as evidenced by their improved mean ranks in the 1,000-sample setting. In contrast, some baseline methods (e.g., IBEA-MS) perform relatively well under limited data but are surpassed by dual-ranking–based GPR methods as the dataset size increases. 

\begin{table}[!h]
\centering
\begin{tabular}{lc}
\toprule
Methods & Overall ranks $\downarrow$\\
\midrule
Prob-RVEA          
& 7.71 $\pm$ 1.27 \\
Prob-MOEA/D        
& 7.21 $\pm$ 1.22 \\
IBEA-MS            
& 3.48 $\pm$ 1.89 \\
TGPR-MO            
& 4.25 $\pm$ 1.94 \\
DDMOEA/GAN         
& 5.92 $\pm$ 3.17 \\
\underline{GPR(RBF)+DR}        
& \underline{2.50 $\pm$ 1.72} \\
\textbf{GPR(Mat\'ern)+DR} 
& \textbf{2.42 $\pm$ 1.84} \\
QR+DR              
& 3.50 $\pm$ 1.56 \\
BNN+DR             
& 6.04 $\pm$ 2.20 \\
\bottomrule
\end{tabular}
\caption{Overall ranking results (mean $\pm$ std.) of different methods, where the surrogate models are trained on a dataset with 1,000 samples. Our method is combined with the \gls{NSGA-II} (omitted) and the dual-ranking (DR) strategy with $\tau=0.9$. \textbf{Bold} indicates the best result, and \underline{underline} indicates the second-best result.}
\label{tab:opt_baseline_1000_ranks}
\end{table}

\section{Conclusion}
In offline data-driven \gls{MOO}, no additional real evaluations can be queried to improve the surrogate model, and uncertainty in surrogate predictions may mislead the optimization process. In this paper, we propose a dual-ranking strategy that jointly leverages surrogate predictions and uncertainty-aware values to mitigate misleading rankings. The prediction experiments motivate our design of the uncertainty-aware function used in dual-ranking. The ablation study demonstrates that the dual-ranking strategy enhances the reliability of surrogate-based evaluations and improves robustness across surrogate models, with limited sensitivity to the target coverage rate or the quantile level $\tau$. A moderate value (e.g., $\tau = 0.90$) provides a well-balanced trade-off between optimization performance and evaluation reliability. Compared with baselines, results show that GPR with a Mat\'ern kernel incorporating the dual-ranking strategy achieves the best overall performance. Moreover, the relatively small standard deviations observed across all dual-ranking–based variants indicate consistently improved robustness.

Future work will include combining the dual-ranking strategy with other optimization algorithms by incorporating their fitness assignment strategies. As the dual-ranking strategy extends the objective dimensionality, we plan to verify the method on many-objective optimization scenarios ($M > 3$). Furthermore, we intend to study real-world cases, such as building space usage optimization problems, as well as online settings with continuously arriving data, where measurement errors exist in real-time sensor data streams.


\bibliographystyle{named}
\bibliography{ijcai26}

\end{document}


\maketitle
\section{Experiments}
\subsection{Experimental Settings}
The \gls{HV} \cite{hypervolume} reference point $(1.1, 1.1)$ and the bounds \cite{benchmark} for Min-Max normalization \cite{indicator} are shown in Table~\ref{tab: reference_points}. 
\begin{table*}[!h]
\centering
\scalebox{0.9}{
\begin{tabular}{lcccc}
\toprule
&Problem &Min &Max &Reference point \\
\midrule
&DTLZ1~\cite{opt_dtlz} &(0,0) &(552.30,568.36) &(1.10, 1.10) \\
&DTLZ2 &(0,0) &(2.78,2.93) &(1.10, 1.10) \\
&DTLZ3 &(0,0) &(1605.54,1670.48) &(1.10, 1.10) \\
&DTLZ4 &(0,0) &(2.83,2.78) &(1.10, 1.10) \\
&DTLZ5 &(0,0) &(2.61,2.70) &(1.10, 1.10) \\
&DTLZ6 &(0,0) &(9.78,9.78) &(1.10, 1.10) \\
&DTLZ7 &(0,0) &(1.10,33.43) &(1.10, 1.10) \\
&BNH~\cite{bnh} &(0,5) &(140,50) &(1.10, 1.10) \\
&Truss2D~\cite{truss2d} &(0,0) &(0.10,1e5) &(1.10, 1.10) \\
&Welded Beam~\cite{welded_beam} &(0,0) &(35,0.10) &(1.10, 1.10) \\
\bottomrule
\end{tabular}
}
\caption{Bounds for Min-Max normalization and hypervolume reference points.}
\label{tab: reference_points}
\end{table*}

\subsection{Surrogate Prediction Performance}
The prediction performance comparison of different surrogate models across 11 benchmark problems with limited training data ($11D-1$ samples) is presented in Table~\ref{tab:prediction}, while the results obtained with 1,000 training samples are shown in Table~\ref{tab:prediction_1000}. We use \gls{MSE} to evaluate prediction accuracy and \gls{CICP}~\cite{cicp} to assess uncertainty estimation performance.

\subsubsection{Prediction Results with Limited Training Data}
From Table~\ref{tab:prediction}, the \gls{GPR} (Matérn) surrogate \cite{kernel_Matern_52} achieves lower MSE than GPR (RBF) surrogate \cite{gaussian_kernel} on many problems, including DTLZ2–DTLZ7 and engineering problems. In contrast, GPR (RBF) tends to achieve slightly higher or comparable \gls{CICP} values on some problems, but often increases the prediction errors. Autogluon-QR \cite{ml_autogluon} consistently delivers high \gls{CICP} values across nearly all cases, demonstrating robust uncertainty estimations despite relatively larger MSE. The BNN \cite{bnn} surrogate exhibits the largest performance variability, with competitive \gls{CICP} values on some problems but degraded prediction accuracy  or \gls{CICP} values on others. These results show a trade-off between prediction accuracy  and reliable uncertainty estimation across all problems.

\begin{table*}[!h]
\centering
\resizebox{\textwidth}{!}{
\begin{tabular}{lcccccccccccc}
\hline
Surrogates
& \multicolumn{2}{c}{DTLZ1}
& \multicolumn{2}{c}{DTLZ2}
& \multicolumn{2}{c}{DTLZ3}
& \multicolumn{2}{c}{DTLZ4}
& \multicolumn{2}{c}{DTLZ5}
& \multicolumn{2}{c}{DTLZ6} \\
\cline{2-13}
& MSE & CICP
& MSE & CICP
& MSE & CICP
& MSE & CICP
& MSE & CICP
& MSE & CICP \\
\hline
GPR (RBF)
& $4.67\mathrm{E}{+03}$ & 88.50\%
& $2.78\mathrm{E}{-02}$ & 76.50\%
& $2.69\mathrm{E}{+04}$ & 87.50\%
& $3.89\mathrm{E}{-02}$ & 84.00\%
& $2.78\mathrm{E}{-02}$ & 76.50\%
& $1.06\mathrm{E}{-02}$ & 53.50\% \\
GPR (Matérn)
& $4.81\mathrm{E}{+03}$ & 88.00\%
& $1.87\mathrm{E}{-02}$ & 73.50\%
& $2.68\mathrm{E}{+04}$ & 87.50\%
& $1.94\mathrm{E}{-02}$ & 90.00\%
& $1.87\mathrm{E}{-04}$ & 73.50\%
& $6.92\mathrm{E}{-03}$ & 74.00\% \\
Autogluon-QR
& $5.01\mathrm{E}{+03}$ & 92.50\%
& $2.45\mathrm{E}{-02}$ & 94.00\%
& $2.89\mathrm{E}{+04}$ & 89.00\%
& $3.21\mathrm{E}{-02}$ & 92.00\%
& $2.45\mathrm{E}{-02}$ & 94.00\%
& $5.97\mathrm{E}{-02}$ & 94.00\% \\
BNN
& $5.00\mathrm{E}{+03}$ & 72.00\%
& $3.01\mathrm{E}{-02}$ & 80.00\%
& $2.83\mathrm{E}{+04}$ & 66.50\%
& $4.43\mathrm{E}{-02}$ & 76.00\%
& $2.99\mathrm{E}{-02}$ & 79.00\%
& $1.39\mathrm{E}{-02}$ & 83.00\% \\
\hline
\multicolumn{13}{c}{} \\
\hline
Surrogates
& \multicolumn{2}{c}{DTLZ7}
& \multicolumn{2}{c}{Omnitest}
& \multicolumn{2}{c}{BNH}
& \multicolumn{2}{c}{Truss2D}
& \multicolumn{2}{c}{Welded Beam}
& \multicolumn{2}{c}{} \\
\cline{2-11}
& MSE & CICP
& MSE & CICP
& MSE & CICP
& MSE & CICP
& MSE & CICP
&  &  \\
\hline
GPR (RBF)
& $4.92\mathrm{E}{-02}$ & 95.50\%
& $8.33\mathrm{E}{-01}$ & 85.00\%
& $1.23\mathrm{E}{-07}$ & 100.00\%
& $3.15\mathrm{E}{+12}$ & 91.00\%
& $4.20\mathrm{E}{+02}$ & 91.00\% \\
GPR (Matérn)
& $3.34\mathrm{E}{-08}$ & 100.00\%
& $8.52\mathrm{E}{-01}$ & 76.00\%
& $2.02\mathrm{E}{-05}$ & 100.00\%
& $3.11\mathrm{E}{+12}$ & 94.00\%
& $4.26\mathrm{E}{+02}$ & 93.00\% \\
Autogluon-QR
& $1.21\mathrm{E}{-01}$ & 91.00\%
& $1.24\mathrm{E}{+00}$ & 78.50\%
& $1.98\mathrm{E}{+01}$ & 79.50\%
& $2.73\mathrm{E}{+12}$ & 97.50\%
& $6.39\mathrm{E}{+02}$ & 91.00\% \\
BNN
& $1.13\mathrm{E}{-01}$ & 65.50\%
& $1.27\mathrm{E}{+00}$ & 53.50\%
& $1.22\mathrm{E}{+01}$ & 95.00\%
& $2.62\mathrm{E}{+12}$ & 96.50\%
& $6.01\mathrm{E}{+02}$ & 92.50\% \\
\hline
\end{tabular}
}
\caption{Prediction performance comparison of different surrogates across 11 benchmark problems.}
\label{tab:prediction}
\end{table*}

\subsubsection{Prediction Results with 1000 Training data}
From Table~\ref{tab:prediction_1000}, all surrogate models benefit from the larger dataset compared to the limited data setting, with substantially reduced MSE and improved \gls{CICP} values across almost all benchmark problems. The GPR (Matérn) surrogate generally achieves lower MSE and higher or comparable CICP values than the GPR (RBF) surrogate, particularly on DTLZ2, DTLZ4–DTLZ7, Truss2D and Welded Beam. Autogluon-QR surrogate maintains relatively high and stable CICP values on most benchmark problems but exhibits larger prediction errors compared to GPR-based models. The BNN surrogate shows improved CICP values on several problems with 1000 training data, however, its prediction accuracy  and CICP values remain more variable than those of other surrogates.

Overall, increasing the training dataset size to 1000 samples improves both prediction accuracy and CICP values for all surrogates. Despite the performance improvement, the relative performance trends among different surrogates remain consistent with those observed under limited training data.

\begin{table*}[!h]
\centering
\resizebox{\textwidth}{!}{
\begin{tabular}{lcccccccccccc}
\hline
Surrogates
& \multicolumn{2}{c}{DTLZ1}
& \multicolumn{2}{c}{DTLZ2}
& \multicolumn{2}{c}{DTLZ3}
& \multicolumn{2}{c}{DTLZ4}
& \multicolumn{2}{c}{DTLZ5}
& \multicolumn{2}{c}{DTLZ6} \\
\cline{2-13}
& MSE & CICP
& MSE & CICP
& MSE & CICP
& MSE & CICP
& MSE & CICP
& MSE & CICP \\
\hline
GPR (RBF) (1000)
& $4.46\mathrm{E}{+03}$ & 88.50\%
& $6.05\mathrm{E}{-09}$ & 98.00\%
& $2.59\mathrm{E}{+04}$ & 88.00\%
& $4.95\mathrm{E}{-04}$ & 88.00\%
& $6.05\mathrm{E}{-09}$ & 98.00\%
& $3.53\mathrm{E}{-03}$ & 82.00\% \\
GPR (Matérn) (1000)
& $4.43\mathrm{E}{+03}$ & 89.00\%
& $5.87\mathrm{E}{-06}$ & 99.50\%
& $2.57\mathrm{E}{+04}$ & 88.00\%
& $5.67\mathrm{E}{-04}$ & 96.00\%
& $5.87\mathrm{E}{-06}$ & 99.50\%
& $2.58\mathrm{E}{-03}$ & 89.00\% \\
Autogluon-QR
& $4.51\mathrm{E}{+03}$ & 86.50\%
& $3.56\mathrm{E}{-03}$ & 90.00\%
& $2.54\mathrm{E}{+04}$ & 85.50\%
& $4.06\mathrm{E}{-03}$ & 91.50\%
& $3.56\mathrm{E}{-03}$ & 90.00\%
& $9.34\mathrm{E}{-03}$ & 96.00\% \\
BNN
& $4.65\mathrm{E}{+03}$ & 75.00\%
& $2.59\mathrm{E}{-02}$ & 83.50\%
& $2.57\mathrm{E}{+04}$ & 64.00\%
& $3.59\mathrm{E}{-02}$ & 77.00\%
& $2.49\mathrm{E}{-02}$ & 74.00\%
& $7.30\mathrm{E}{-03}$ & 96.50\% \\
\hline
\multicolumn{13}{c}{} \\
\hline
Surrogates
& \multicolumn{2}{c}{DTLZ7}
& \multicolumn{2}{c}{Omnitest}
& \multicolumn{2}{c}{BNH}
& \multicolumn{2}{c}{Truss2D}
& \multicolumn{2}{c}{Welded Beam}
& \multicolumn{2}{c}{} \\
\cline{2-11}
& MSE & CICP
& MSE & CICP
& MSE & CICP
& MSE & CICP
& MSE & CICP
&  &  \\
\hline
GPR (RBF) (1000)
& $2.76\mathrm{E}{-10}$ & 100.00\%
& $2.19\mathrm{E}{-10}$ & 100.00\%
& $1.78\mathrm{E}{-07}$ & 100.00\%
& $2.74\mathrm{E}{+12}$ & 99.50\%
& $9.41\mathrm{E}{+01}$ & 98.00\% \\
GPR (Matérn) (1000)
& $2.48\mathrm{E}{-10}$ & 99.50\%
& $2.26\mathrm{E}{-06}$ & 100.00\%
& $4.74\mathrm{E}{-07}$ & 100.00\%
& $2.46\mathrm{E}{+12}$ & 98.00\%
& $9.44\mathrm{E}{+00}$ & 98.50\% \\
Autogluon-QR
& $4.97\mathrm{E}{-02}$ & 90.00\%
& $5.12\mathrm{E}{-03}$ & 91.00\%
& $2.25\mathrm{E}{-01}$ & 98.50\%
& $1.19\mathrm{E}{+12}$ & 97.50\%
& $8.24\mathrm{E}{+01}$ & 52.50\% \\
BNN
& $7.49\mathrm{E}{-02}$ & 88.00\%
& $1.52\mathrm{E}{-03}$ & 70.00\%
& $5.49\mathrm{E}{+00}$ & 99.50\%
& $2.73\mathrm{E}{+12}$ & 96.50\%
& $3.40\mathrm{E}{+01}$ & 99.00\% \\
\hline
\end{tabular}
}
\caption{Prediction performance comparison of different surrogates trained on 1000 samples across 11 benchmark problems.}
\label{tab:prediction_1000}
\end{table*}

\subsection{Ablation Study and Sensitivity Analysis}
The optimization performance indicators include \gls{MSE}, \gls{IGD+}~\cite{igd_plus}, and \gls{HV}~\cite{hypervolume}. For both \gls{MSE} and \gls{IGD+}, smaller values indicate better performance, whereas larger \gls{HV} values are preferred. As all surrogates are combined with the same optimization algorithm \gls{NSGA-II}, this is omitted in the table.

\subsubsection{Ablation Study and Sensitivity Analysis with Limited Training Data}
The ablation study results of optimization performance comparisons and sensitivity analysis with limited training data ($11D-1$ samples) are presented in Table~\ref{tab:opt}. The dual-ranking strategy consistently improves or preserves optimization performance for all surrogate models combined with \gls{NSGA-II}. These improvements are not limited to a specific problem but are observed across both unconstrained and constrained benchmark problems.

For the GPR (RBF) method with dual-ranking strategy, MSE is reduced on all DTLZ problems and on 3 out of 4 remaining problems, while performance remains unchanged or only marginally affected in the other cases. IGD+ is improved in 4 out of 7 DTLZ problems and in 3 out of 4 other problems, with no severe degradation observed. The HV metric is largely preserved across DTLZ problems and significantly improved in 2 out of 4 other problems, particularly on constrained benchmarks such as BNH and Truss2D. Overall, these results indicate that dual-ranking strategy substantially enhances solution quality for the GPR (RBF) method.

For GPR with the Matérn kernel, the baseline model already exhibits strong performance on several problems. Dual-ranking strategy further improves optimization outcomes. Across all benchmark problems, both MSE and IGD+ are reduced in 6 out of 11 cases, while HV performance is improved or preserved in many problems. In particular, constrained benchmarks benefit the most from the dual-ranking strategy, where \gls{MSE} and \gls{IGD+} are consistently reduced while \gls{HV} is improved. This demonstrates that dual-ranking strategy remains effective even when combined with a well-performing surrogate model.

For the Autogluon-QR method, the dual-ranking strategy produces consistent improvements across almost all benchmark problems. MSE is reduced in 7 out of 11 cases and \gls{IGD+} is reduced in 8 out of 11 cases. HV is improved or preserved in at least 10 problems. These results show that dual-ranking strategy effectively improves or preserves the optimization performance.

The BNN model exhibits the largest performance variability in the baseline setting. After incorporating dual-ranking strategy, optimization results become substantially more stable. On unconstrained problems, dual-ranking strategy improves IGD+ and HV in most cases, even when MSE improvements are limited. For constrained problems, dual-ranking strategy preserves or improves solution quality. This indicates enhanced robustness across both unconstrained and constrained scenarios.

Across all methods, these improvement patterns remain consistent under different dual-ranking parameter settings. No single parameter choice leads to performance degradation across all problems. In most cases, intermediate to large parameter values (e.g., $\tau=0.90$ for GPR-based models and $\tau=0.90$ for QR and BNN) achieve a good balance. Overall, the results confirm that the proposed dual-ranking strategy delivers robust performance gains across diverse problems without requiring careful hyperparameter tuning.

\begin{table*}[!h]
\centering
\resizebox{\textwidth}{!}{
\begin{tabular}{l ccc ccc ccc ccc ccc ccc}
\hline
Methods
& \multicolumn{3}{c}{DTLZ1}
& \multicolumn{3}{c}{DTLZ2}
& \multicolumn{3}{c}{DTLZ3}
& \multicolumn{3}{c}{DTLZ4}
& \multicolumn{3}{c}{DTLZ5}
& \multicolumn{3}{c}{DTLZ6} \\
\cmidrule(lr){2-4} \cmidrule(lr){5-7} \cmidrule(lr){8-10} \cmidrule(lr){11-13} \cmidrule(lr){14-16} \cmidrule(lr){17-19}
& MSE & IGD+ & HV
& MSE & IGD+ & HV
& MSE & IGD+ & HV
& MSE & IGD+ & HV
& MSE & IGD+ & HV
& MSE & IGD+ & HV \\
\hline
GPR (RBF)
& 1.01E+04 & 1.08E+02 & 1.15
& 5.92E-02 & 5.49E-01 & 0.96
& 2.53E+04 & 4.25E+02 & 1.11
& 2.70E-01 & 4.50E-01 & 0.85
& 5.92E-02 & 5.43E-01 & 0.92
& 1.52E+00 & 4.40E+00 & 0.95 \\
GPR + DR ($\tau=0.80$)
& \textbf{9.74E+03} & 1.11E+02 & \underline{1.15}
& \textbf{5.27E-02} & \textbf{5.32E-01} & \underline{0.96}
& 2.58E+04 & \textbf{3.92E+02} & \textbf{1.12}
& \textbf{1.98E-01} & 4.56E-01 & 0.83
& \textbf{5.27E-02} & \textbf{5.28E-01} & \underline{0.92}
& \textbf{1.06E+00} & 4.65E+00 & 0.93 \\
GPR + DR ($\tau=0.90$)
& \textbf{9.43E+03} & 1.16E+02 & 1.14
& \textbf{5.44E-02} & \textbf{5.17E-01} & \textbf{0.97}
& \textbf{2.28E+04} & 4.18E+02 & \underline{1.11}
& \textbf{2.08E-01} & 4.73E-01 & 0.83
& \textbf{5.44E-02} & \textbf{5.12E-01} & \textbf{0.93}
& \textbf{6.32E-01} & 4.98E+00 & 0.89 \\
GPR + DR ($\tau=0.95$)
& \textbf{8.36E+03} & 1.25E+02 & 1.13
& \textbf{5.11E-02} & \textbf{5.21E-01} & \textbf{0.97}
& 2.57E+04 & \textbf{4.12E+02} & \textbf{1.12}
& \textbf{1.78E-01} & 4.59E-01 & 0.82
& \textbf{5.11E-02} & \textbf{5.19E-01} & \textbf{0.93}
& \textbf{3.68E-01} & 5.29E+00 & 0.86 \\
\hline
GPR (Matérn)
& 9.59E+03 & 1.19E+02 & 1.14
& 1.54E-02 & 8.32E-02 & 1.09
& 2.54E+04 & 3.97E+02 & 1.12
& 1.85E-01 & 4.64E-01 & 0.79
& 1.54E-02 & 8.24E-02 & 1.08
& 2.37E+00 & 4.12E+00 & 0.98 \\
GPR + DR ($\tau=0.80$)
& \textbf{8.44E+03} & 1.25E+02 & 1.13
& 1.77E-02 & \textbf{7.35E-02} & \textbf{1.10}
& \textbf{2.38E+04} & 4.19E+02 & 1.11
& 2.15E-01 & 4.65E-01 & \textbf{0.80}
& 1.77E-02 & \textbf{7.35E-02} & \underline{1.08}
& \textbf{1.81E+00} & 4.36E+00 & 0.95 \\
GPR + DR ($\tau=0.90$)
& \textbf{7.94E+03} & 1.33E+02 & 1.12
& 2.00E-02 & \textbf{6.61E-02} & \textbf{1.10}
& \textbf{2.27E+04} & 4.24E+02 & 1.11
& 2.43E-01 & \textbf{4.43E-01} & \textbf{0.80}
& 2.00E-02 & \textbf{6.68E-02} & \underline{1.08}
& \textbf{1.45E+00} & 4.51E+00 & 0.94 \\
GPR + DR ($\tau=0.95$)
& \textbf{7.61E+03} & 1.24E+02 & 1.13
& 1.95E-02 & \textbf{6.15E-02} & \textbf{1.10}
& \textbf{2.24E+04} & 4.29E+02 & 1.11
& 2.35E-01 & \textbf{4.21E-01} & \textbf{0.80}
& 1.95E-02 & \textbf{6.21E-02} & \underline{1.08}
& \textbf{1.17E+00} & 4.68E+00 & 0.92 \\
\hline
Autogluon (QR)
& 8.06E+03 & 1.85E+02 & 1.04
& 2.15E-01 & 2.00E-01 & 1.06
& 3.45E+04 & 4.51E+02 & 1.11
& 9.47E-01 & 5.73E-01 & 0.88
& 2.15E-01 & 2.08E-01 & 1.03
& 1.97E-01 & 6.70E+00 & 0.68 \\
QR + DR ($q=0.80$)
& \textbf{6.86E+03} & 1.93E+02 & 1.02
& \textbf{5.95E-02} & \textbf{1.40E-01} & \textbf{1.07}
& 3.91E+04 & \textbf{4.46E+02} & 1.10
& \textbf{5.61E-01} & \textbf{3.24E-01} & \textbf{0.93}
& \textbf{5.95E-02} & \textbf{1.44E-01} & \textbf{1.05}
& 2.28E-01 & \textbf{6.61E+00} & \textbf{0.69} \\
QR + DR ($\tau=0.90$)
& \textbf{6.63E+03} & 2.06E+02 & 1.01
& \textbf{5.23E-02} & \textbf{1.16E-01} & \textbf{1.08}
& 3.85E+04 & 4.89E+02 & 1.10
& \textbf{6.00E-01} & \textbf{1.96E-01} & \textbf{1.01}
& \textbf{5.23E-02} & \textbf{1.18E-01} & \textbf{1.06}
& 2.15E-01 & \textbf{6.66E+00} & \underline{0.68} \\
QR + DR ($\tau=0.95$)
& \textbf{6.69E+03} & 2.08E+02 & 1.01
& \textbf{4.94E-02} & \textbf{1.23E-01} & \textbf{1.08}
& 3.47E+04 & 4.62E+02 & \underline{1.11}
& \textbf{5.17E-01} & \textbf{2.01E-01} & \textbf{1.00}
& \textbf{4.94E-02} & \textbf{1.25E-01} & \textbf{1.06}
& 2.12E-01 & 6.73E+00 & \underline{0.68} \\
\hline
BNN
& 5.56E+03 & 1.96E+02 & 1.02
& 6.73E-01 & 1.14E+00 & 0.69
& 4.09E+04 & 5.26E+02 & 1.08
& 7.95E-01 & 1.53E+00 & 0.43
& 5.43E-01 & 1.22E+00 & 0.59
& 9.66E-01 & 5.38E+00 & 0.84 \\
BNN + DR ($\tau=0.80$)
& 7.68E+03 & 2.03E+02 & 1.01
& \textbf{6.04E-01} & \textbf{1.11E+00} & \textbf{0.71}
& \textbf{3.50E+04} & 5.43E+02 & \underline{1.08}
& \textbf{6.95E-01} & \textbf{1.47E+00} & \textbf{0.46}
& \textbf{5.01E-01} & \textbf{1.18E+00} & \textbf{0.61}
& \textbf{8.29E-01} & 5.53E+00 & 0.83 \\
BNN + DR ($\tau=0.90$)
& 8.72E+03 & 2.25E+02 & 0.98
& \textbf{5.72E-01} & \textbf{1.10E+00} & \textbf{0.72}
& \textbf{3.70E+04} & \underline{5.26E+02} & \underline{1.08}
& \textbf{6.67E-01} & \textbf{1.39E+00} & \textbf{0.50}
& \textbf{4.97E-01} & \textbf{1.15E+00} & \textbf{0.63}
& \textbf{9.03E-01} & 5.48E+00 & 0.83 \\
BNN + DR ($\tau=0.95$)
& 6.61E+03 & 2.13E+02 & 1.00
& \textbf{5.49E-01} & \textbf{1.04E+00} & \textbf{0.75}
& \textbf{3.88E+04} & 5.46E+02 & \underline{1.08}
& \textbf{6.72E-01} & \textbf{1.38E+00} & \textbf{0.50}
& \textbf{4.69E-01} & \textbf{1.11E+00} & \textbf{0.64}
& \textbf{8.69E-01} & 5.49E+00 & 0.83 \\
\hline
\multicolumn{19}{c}{} \\
\hline
Methods
& \multicolumn{3}{c}{DTLZ7}
& \multicolumn{3}{c}{Omnitest}
& \multicolumn{3}{c}{BNH}
& \multicolumn{3}{c}{Truss2D}
& \multicolumn{3}{c}{Welded Beam}
& \multicolumn{3}{c}{} \\
\cmidrule(lr){2-4} \cmidrule(lr){5-7} \cmidrule(lr){8-10} \cmidrule(lr){11-13} \cmidrule(lr){14-16}
& MSE & IGD+ & HV
& MSE & IGD+ & HV
& MSE & IGD+ & HV
& MSE & IGD+ & HV
& MSE & IGD+ & HV
&  &  &  \\
\hline
GPR (RBF)
& 1.29E-01 & 1.11E-01 & 1.10
& 3.44E-01 & 2.95E-01 & 1.01
& 3.75E-07 & 1.91E-01 & 1.05
& 1.17E+11 & 1.40E+04 & 0.67
& 2.72E+00 & 3.98E-01 & 1.11 \\
GPR + DR ($\tau=0.80$)
& \textbf{1.15E-01} & 1.30E-01 & \underline{1.10}
& 4.56E-01 & \textbf{2.19E-01} & \textbf{1.04}
& \textbf{3.73E-07} & \textbf{1.88E-01} & \underline{1.05}
& \textbf{9.97E+10} & \textbf{7.32E+02} & \textbf{0.95}
& 2.74E+00 & 4.01E-01 & \underline{1.11} \\
GPR + DR ($\tau=0.90$)
& \textbf{1.13E-01} & 1.30E-01 & \underline{1.10}
& 4.37E-01 & \textbf{2.15E-01} & \textbf{1.06}
& 3.88E-07 & \textbf{1.88E-01} & \underline{1.05}
& \textbf{9.04E+10} & \textbf{6.98E+02} & \textbf{0.95}
& \textbf{2.42E+00} & 4.22E-01 & \underline{1.11} \\
GPR + DR ($\tau=0.95$)
& \textbf{1.07E-01} & \textbf{9.60E-02} & \underline{1.10}
& 4.25E-01 & \textbf{2.09E-01} & \textbf{1.07}
& 3.81E-07 & \textbf{1.86E-01} & \underline{1.05}
& \textbf{6.19E+10} & \textbf{5.03E+02} & \textbf{0.95}
& 2.90E+00 & \textbf{3.91E-01} & \textbf{1.13} \\
\hline
GPR (Matérn)
& 1.17E-07 & 3.76E-03 & 1.11
& 4.70E-01 & 2.33E-01 & 1.03
& 1.28E-04 & 1.85E-01 & 1.05
& 5.34E+10 & 6.74E+03 & 0.87
& 2.54E+00 & 4.60E-01 & 1.06 \\
GPR + DR ($\tau=0.80$)
& 1.25E-07 & 4.49E-03 & \underline{1.11}
& 5.46E-01 & \textbf{1.96E-01} & \textbf{1.07}
& \textbf{1.25E-04} & 1.89E-01 & \underline{1.05}
& \textbf{5.14E+10} & \textbf{1.12E+03} & \textbf{0.95}
& \textbf{1.51E+00} & \textbf{4.22E-01} & \textbf{1.10} \\
GPR + DR ($\tau=0.90$)
& 1.41E-07 & 4.21E-03 & \underline{1.11}
& 5.56E-01 & \textbf{1.95E-01} & \textbf{1.08}
& \textbf{1.25E-04} & 1.86E-01 & \underline{1.05}
& \textbf{4.24E+10} & \textbf{7.01E+02} & \textbf{0.95}
& \textbf{1.52E+00} & \textbf{3.76E-01} & \textbf{1.13} \\
GPR + DR ($\tau=0.95$)
& 1.32E-07 & 4.08E-03 & \underline{1.11}
& 5.35E-01 & \textbf{2.00E-01} & \textbf{1.08}
& \textbf{1.25E-04} & 1.87E-01 & \underline{1.05}
& \textbf{2.86E+10} & \textbf{5.54E+02} & \textbf{0.96}
& \textbf{1.86E+00} & 6.06E-01 & \textbf{1.10} \\
\hline
Autogluon (QR)
& 3.27E+00 & 9.98E-02 & 1.10
& 5.72E-01 & 1.17E+00 & 0.70
& 2.62E+01 & 4.90E-01 & 1.04
& 7.48E+08 & 9.83E-03 & 0.94
& 8.89E+01 & 5.93E+00 & 1.10 \\
QR + DR ($\tau=0.80$)
& 3.49E+00 & 1.03E-01 & \underline{1.10}
& \textbf{2.95E-01} & \textbf{4.60E-01} & \textbf{0.93}
& 2.67E+01 & 5.63E-01 & \underline{1.04}
& \textbf{6.37E+08} & 2.36E-02 & \textbf{0.96}
& 1.12E+02 & \textbf{5.08E+00} & \textbf{1.11} \\
QR + DR ($\tau=0.90$)
& 3.56E+00 & \textbf{8.00E-02} & \underline{1.10}
& \textbf{3.70E-01} & \textbf{5.00E-01} & \textbf{0.91}
& \textbf{2.50E+01} & 5.31E-01 & \underline{1.04}
& \textbf{6.35E+08} & 1.47E-02 & \textbf{0.97}
& 1.09E+02 & 6.22E+00 & \underline{1.10} \\
QR + DR ($\tau=0.95$)
& 3.68E+00 & \textbf{8.82E-02} & \underline{1.10}
& \textbf{4.15E-01} & \textbf{5.75E-01} & \textbf{0.89}
& \textbf{2.59E+01} & 5.36E-01 & \underline{1.04}
& \textbf{6.70E+08} & 1.42E-02 & \textbf{0.96}
& 9.19E+01 & \textbf{4.72E+00} & \textbf{1.12} \\
\hline
BNN
& 2.30E+00 & 1.23E-01 & 1.10
& 1.77E+00 & 1.93E-01 & 1.11
& 7.11E+00 & 2.65E+01 & 0.54
& 1.58E+13 & 1.05E+06 & 0.58
& 1.17E+02 & 4.98E+01 & 0.41 \\
BNN + DR ($\tau=0.80$)
& \textbf{1.96E+00} & 1.78E-01 & \underline{1.10}
& 1.78E+00 & \textbf{1.52E-01} & \textbf{1.12}
& 9.23E+00 & \underline{2.65E+01} & \underline{0.54}
& \underline{1.58E+13} & \underline{1.05E+06} & \underline{0.58}
& 1.35E+02 & \underline{4.98E+01} & \underline{0.41} \\
BNN + DR ($\tau=0.90$)
& \textbf{1.85E+00} & 1.66E-01 & \underline{1.10}
& 1.78E+00 & \textbf{1.47E-01} & \textbf{1.12}
& 7.36E+00 & \underline{2.65E+01} & \underline{0.54}
& \underline{1.58E+13} & \underline{1.05E+06} & \underline{0.58}
& 1.49E+02 & \underline{4.98E+01} & \underline{0.41} \\
BNN + DR ($\tau=0.95$)
& \textbf{1.81E+00} & 1.96E-01 & 1.09
& \textbf{1.72E+00} & \textbf{1.48E-01} & \textbf{1.12}
& 8.75E+00 & \underline{2.65E+01} & \underline{0.54}
& \underline{1.58E+13} & \underline{1.05E+06} & \underline{0.58}
& 1.29E+02 & \underline{4.98E+01} & \underline{0.41} \\
\hline
\end{tabular}
}
\caption{Ablation study and sensitivity analysis of optimization performance comparisons across 11 benchmark problems. Each method is evaluated under different Dual-Ranking (DR) settings (the target coverage rate $\tau = 0.80, 0.90, 0.95$ or the quantile level $\tau = 0.80, 0.90, 0.95$) for sensitivity analysis. \textbf{Bold} indicates improvement over the baseline without DR, and \underline{underline} indicates equal performance.}
\label{tab:opt}
\end{table*}

\subsubsection{Ablation Study Between Dual-Ranking and the Uncertainty-Aware Function}
We summarize the findings of the ablation study comparing surrogates combined with the dual-ranking strategy and those combined with the uncertainty-aware function $f_{\text{UA-sur}}$, as shown in Tables~\ref{tab:opt_sur} and~\ref{tab:opt_sur_counts}.

Overall, the dual-ranking strategy achieves more improvements (or ties) in IGD+ and HV than in MSE. The dual-ranking strategy may retain solutions that are promising under real evaluations but lie in regions with higher surrogate prediction errors, resulting in larger MSE values while still improving IGD+ and HV. Consequently, the dual-ranking strategy can improve the final optimization performance.

\begin{table*}[!h]
\centering
\resizebox{\textwidth}{!}{
\begin{tabular}{l ccc ccc ccc ccc ccc ccc}
\toprule
\multirow{2}{*}{Methods}
& \multicolumn{3}{c}{DTLZ1}
& \multicolumn{3}{c}{DTLZ2}
& \multicolumn{3}{c}{DTLZ3}
& \multicolumn{3}{c}{DTLZ4}
& \multicolumn{3}{c}{DTLZ5}
& \multicolumn{3}{c}{DTLZ6} \\
\cmidrule(lr){2-4} \cmidrule(lr){5-7} \cmidrule(lr){8-10}
\cmidrule(lr){11-13} \cmidrule(lr){14-16} \cmidrule(lr){17-19}
& MSE & IGD+ & HV
& MSE & IGD+ & HV
& MSE & IGD+ & HV
& MSE & IGD+ & HV
& MSE & IGD+ & HV
& MSE & IGD+ & HV \\
\midrule

GPR (RBF) + DR
& 9.43E+03 & 1.16E+02 & \underline{1.14}
& 5.44E-02 & \textbf{5.17E-01} & \textbf{0.97}
& \textbf{2.28E+04} & 4.18E+02 & 1.11
& 2.08E-01 & \textbf{4.73E-01} & \textbf{0.83}
& 5.44E-02 & \textbf{5.12E-01} & \textbf{0.93}
& 6.32E-01 & \textbf{4.98E+00} & \underline{0.89} \\
GPR (RBF) + $f_{\text{UA-sur}}$
& 9.36E+03 & 1.10E+02 & 1.14
& 5.31E-02 & 5.37E-01 & 0.96
& 2.40E+04 & 3.95E+02 & 1.12
& 1.22E-01 & 4.84E-01 & 0.80
& 5.31E-02 & 5.33E-01 & 0.92
& 6.04E-01 & 5.03E+00 & 0.89 \\

\hline
GPR (Mat\'ern) + DR
& 7.94E+03 & \textbf{1.33E+02} & \underline{1.12}
& 2.00E-02 & 6.61E-02 & \underline{1.10}
& \textbf{2.27E+04} & 4.24E+02 & 1.11
& 2.43E-01 & 4.43E-01 & 0.80
& 2.00E-02 & 6.68E-02 & \underline{1.08}
& 1.45E+00 & \textbf{4.51E+00} & \textbf{0.94} \\
GPR (Mat\'ern) + $f_{\text{UA-sur}}$
& 7.02E+03 & 1.34E+02 & 1.12
& 1.97E-02 & 6.23E-02 & 1.10
& 2.44E+04 & 4.05E+02 & 1.12
& 9.50E-02 & 4.07E-01 & 0.81
& 1.97E-02 & 6.26E-02 & 1.08
& 1.35E+00 & 4.60E+00 & 0.93 \\

\hline
QR + DR
& 6.63E+03 & 2.06E+02 & \underline{1.01}
& \textbf{5.23E-02} & 1.16E-01 & \underline{1.08}
& 3.85E+04 & 4.89E+02 & \underline{1.10}
& 6.00E-01 & \textbf{1.96E-01} & \textbf{1.01}
& \textbf{5.23E-02} & 1.18E-01 & \underline{1.06}
& 2.15E-01 & \textbf{6.66E+00} & \textbf{0.68} \\
QR + $f_{\text{UA-sur}}$
& 6.22E+03 & 1.97E+02 & 1.01
& 5.71E-02 & 1.10E-01 & 1.08
& 3.30E+04 & 4.42E+02 & 1.10
& 2.29E-01 & 5.37E-01 & 0.76
& 5.71E-02 & 1.10E-01 & 1.06
& 2.08E-01 & 6.75E+00 & 0.67 \\

\hline
BNN + DR
& \textbf{8.72E+03} & 2.25E+02 & \textbf{0.98}
& 5.72E-01 & 1.10E+00 & 0.72
& \textbf{3.70E+04} & \textbf{5.26E+02} & \textbf{1.08}
& 6.67E-01 & 1.39E+00 & 0.50
& 4.97E-01 & 1.15E+00 & 0.63
& 9.03E-01 & \textbf{5.48E+00} & \textbf{0.83} \\
BNN + $f_{\text{UA-sur}}$
& 1.37E+04 & 2.22E+02 & 0.97
& 3.33E-01 & 9.53E-01 & 0.79
& 5.92E+04 & 5.41E+02 & 1.06
& 4.45E-01 & 1.31E+00 & 0.51
& 4.88E-01 & 1.05E+00 & 0.67
& 3.75E-01 & 6.33E+00 & 0.73 \\

\hline
\multicolumn{19}{c}{} \\
\hline
Methods
& \multicolumn{3}{c}{DTLZ7}
& \multicolumn{3}{c}{Omnitest}
& \multicolumn{3}{c}{BNH}
& \multicolumn{3}{c}{Truss2D}
& \multicolumn{3}{c}{Welded Beam} \\
\cmidrule(lr){2-4} \cmidrule(lr){5-7}
\cmidrule(lr){8-10} \cmidrule(lr){11-13} \cmidrule(lr){14-16}
& MSE & IGD+ & HV
& MSE & IGD+ & HV
& MSE & IGD+ & HV
& MSE & IGD+ & HV
& MSE & IGD+ & HV \\
\midrule

GPR (RBF) + DR
& 1.13E-01 & 1.30E-01 & \underline{1.10}
& 4.37E-01 & \textbf{2.15E-01} & \textbf{1.06}
& \textbf{3.88E-07} & \textbf{1.88E-01} & \underline{1.05}
& 9.04E+10 & 6.98E+02 & 0.95
& \textbf{2.42E+00} & 4.22E-01 & \textbf{1.11} \\
GPR (RBF) + $f_{\text{UA-sur}}$
& 1.11E-01 & 1.12E-01 & 1.10
& 2.43E-01 & 2.68E-01 & 1.03
& 3.89E-07 & 1.95E-01 & 1.05
& 2.05E+10 & 2.18E+00 & 0.96
& 3.10E+00 & 3.04E-01 & 1.10 \\

\hline
GPR (Mat\'ern) + DR
& 1.41E-07 & 4.21E-03 & \underline{1.11}
& 5.56E-01 & \textbf{1.95E-01} & \textbf{1.08}
& \textbf{1.25E-04} & \textbf{1.86E-01} & \underline{1.05}
& 4.24E+10 & \textbf{7.01E+02} & \textbf{0.95}
& \textbf{1.52E+00} & 3.76E-01 & \textbf{1.13} \\
GPR (Mat\'ern) + $f_{\text{UA-sur}}$
& 1.29E-07 & 3.76E-03 & 1.11
& 1.56E-01 & 3.04E-01 & 1.01
& 1.27E-04 & 1.87E-01 & 1.05
& 3.77E+10 & 3.95E+03 & 0.90
& 3.29E+00 & 3.27E-01 & 1.10 \\

\hline
QR + DR
& \textbf{3.56E+00} & 8.00E-02 & \underline{1.10}
& 3.70E-01 & 5.00E-01 & 0.91
& \textbf{2.50E+01} & \textbf{5.31E-01} & \underline{1.04}
& 6.35E+08 & \textbf{1.47E-02} & \textbf{0.97}
& \textbf{1.09E+02} & \textbf{6.22E+00} & \textbf{1.10} \\
QR + $f_{\text{UA-sur}}$
& 4.90E+00 & 2.71E-02 & 1.10
& 3.19E-01 & 4.55E-01 & 0.93
& 2.64E+01 & 6.47E-01 & 1.04
& 6.00E+08 & 1.77E-02 & 0.92
& 2.12E+02 & 1.54E+01 & 1.01 \\

\hline
BNN + DR
& \underline{1.85E+00} & \textbf{1.66E-01} & \textbf{1.10}
& \textbf{1.78E+00} & \textbf{1.47E-01} & \textbf{1.12}
& 7.36E+00 & \underline{2.65E+01} & \underline{0.54}
& \textbf{1.58E+13} & \underline{1.05E+06} & \underline{0.58}
& 1.49E+02 & \underline{4.98E+01} & \underline{0.41} \\
BNN + $f_{\text{UA-sur}}$
& 1.85E+00 & 3.11E-01 & 1.09
& 2.12E+00 & 4.04E-01 & 1.03
& 7.21E+00 & 2.65E+01 & 0.54
& 1.59E+13 & 1.05E+06 & 0.58
& 1.48E+02 & 4.98E+01 & 0.41 \\

\bottomrule
\end{tabular}
}
\caption{Ablation study of optimization performance comparisons between surrogates combined with dual-ranking (DR) and those combined with $f_{\text{UA-sur}}$. \textbf{Bold} indicates better performance, and \underline{underline} indicates equal performance.}
\label{tab:opt_sur}
\end{table*}

\begin{table}[!htbp]
\centering
\setlength{\tabcolsep}{3pt}
\begin{tabular}{l|ccc|ccc|ccc}
\hline
\multirow{2}{*}{Methods} 
& \multicolumn{3}{c|}{MSE} 
& \multicolumn{3}{c|}{IGD+} 
& \multicolumn{3}{c}{HV} \\
 & $+$ & $=$ & $-$ & $+$ & $=$ & $-$ & $+$ & $=$ & $-$ \\
\hline
GPR (RBF) + DR      & 3 & 0 & 8 & 6 & 0 & 5 & 5 & 4 & 2 \\
GPR (Mat\'ern) + DR & 3 & 0 & 8 & 5 & 0 & 6 & 4 & 5 & 2 \\
QR + DR             & 5 & 0 & 6 & 5 & 0 & 6 & 4 & 6 & 1 \\
BNN + DR            & 4 & 1 & 6 & 4 & 3 & 4 & 5 & 3 & 3 \\
\hline
\end{tabular}
\caption{Counts based on Table \ref{tab:opt_sur}. For each indicator, `$+$/$=$/$-$'' indicates better/equal/worse performance of surrogates combined with dual-ranking (DR) over those combined with $f_{\text{UA-sur}}$.}
\label{tab:opt_sur_counts}
\end{table}

\subsubsection{Optimization Results with 1000 Training data}
The optimization results obtained with 1,000 training samples are shown in Table~\ref{tab:opt_1000}. We take the hyperparameter $\tau=0.9$ and test the methods where the surrogates trained on the dataset with 1000 samples. As shown in Table \ref{tab:opt_1000}, the proposed dual-ranking strategy provides consistent benefits for different methods across most problems.

For GPR (RBF) method, introducing dual-ranking strategy improves the MSE on 7 out of 11 problems and leads to better IGD+ on some benchmarks, while HV remains unchanged in 9 out of 11 cases. The performance degradation on a few problems such as DTLZ7 and Welded Beam. For the GPR (Matérn) method, the baseline already exhibits strong optimization performance. \gls{HV} values remain largely unchanged, while improvements in \gls{IGD+} are limited in many cases. For AutoGluon-QR method, the dual-ranking strategy substantially reduces \gls{MSE} on most benchmark problems. However, improvements in \gls{IGD+} and \gls{HV} are observed only on a subset of benchmarks. For BNN method, baseline optimization performance shows large variability across problems. Incorporating dual-ranking strategy improves or preserves MSE and HV on multiple benchmarks, although performance degradation is still observed on a few problems. 

In summary, all methods benefit from the dual-ranking strategy, particularly in terms of \gls{MSE}, indicating that the surrogate-based evaluations of the obtained solutions become closer to the real evaluations. In contrast, the effectiveness of the dual-ranking strategy on \gls{IGD+} is more dependent on the surrogate models and problems. For GPR (RBF) and BNN, improvements in \gls{IGD+} are relatively balanced, with several cases exhibiting comparable performance. However, for GPR (Matérn) and AutoGluon-QR, reductions in \gls{IGD+} are observed on the majority of problems, suggesting that the dual-ranking strategy does not consistently guarantee improved convergence. This may also be attributed to limitations in uncertainty estimation quality, which can reduce the effectiveness of dual-ranking in some cases.

\begin{table*}[htbp]
\centering
\resizebox{\textwidth}{!}{
\begin{tabular}{l ccc ccc ccc ccc ccc ccc}
\toprule
\multirow{2}{*}{Methods}
& \multicolumn{3}{c}{DTLZ1}
& \multicolumn{3}{c}{DTLZ2}
& \multicolumn{3}{c}{DTLZ3}
& \multicolumn{3}{c}{DTLZ4}
& \multicolumn{3}{c}{DTLZ5}
& \multicolumn{3}{c}{DTLZ6} \\
\cmidrule(lr){2-4} \cmidrule(lr){5-7} \cmidrule(lr){8-10} \cmidrule(lr){11-13} \cmidrule(lr){14-16} \cmidrule(lr){17-19}
& MSE & IGD+ & HV
& MSE & IGD+ & HV
& MSE & IGD+ & HV
& MSE & IGD+ & HV
& MSE & IGD+ & HV
& MSE & IGD+ & HV \\
\midrule
GPR (RBF) (1000)
& 9.78E+03 & 1.41E+02 & 1.12
& 1.85E-08 & 2.58E-03 & 1.11
& 3.09E+04 & 4.16E+02 & 1.12
& 2.43E-01 & 3.08E-01 & 0.99
& 1.85E-08 & 2.75E-03 & 1.10
& 1.49E+00 & 4.35E+00 & 0.95 \\
GPR + DR ($\tau=0.90$)
& 1.21E+04 & \textbf{1.34E+02} & \underline{1.12}
& \textbf{1.81E-08} & \underline{2.58E-03} & \underline{1.11}
& \textbf{2.86E+04} & \textbf{4.05E+02} & \underline{1.12}
& \textbf{2.06E-01} & \textbf{2.76E-01} & \textbf{1.02}
& \textbf{1.81E-08} & \textbf{2.74E-03} & \underline{1.10}
& \textbf{9.51E-01} & 4.66E+00 & 0.92 \\
\hline
GPR (Matérn) (1000)
& 1.66E+04 & 1.27E+02 & 1.14
& 4.38E-05 & 3.07E-03 & 1.11
& 1.17E+05 & 2.95E+02 & 1.17
& 3.98E-02 & 1.20E-01 & 1.08
& 4.38E-05 & 3.27E-03 & 1.10
& 2.49E+00 & 3.97E+00 & 0.99 \\
GPR + DR ($\tau=0.90$)
& \textbf{1.49E+04} & 1.32E+02 & 1.13
& 4.51E-05 & 3.09E-03 & \underline{1.11}
& \textbf{1.09E+05} & 3.23E+02 & 1.16
& 4.28E-02 & \textbf{1.09E-01} & \underline{1.08}
& 4.51E-05 & 3.32E-03 & \underline{1.10}
& \textbf{2.45E+00} & 3.99E+00 & \underline{0.99} \\
\hline
Autogluon (QR) (1000)
& 6.04E+03 & 1.04E+02 & 1.15
& 5.01E-03 & 1.01E-02 & 1.11
& 3.74E+04 & 2.38E+02 & 1.18
& 1.78E-03 & 3.58E-01 & 0.84
& 5.01E-03 & 1.01E-02 & 1.10
& 1.68E+00 & 5.03E+00 & 0.88 \\
QR + DR ($\tau=0.90$)
& \textbf{4.25E+03} & 1.34E+02 & 1.11
& 5.43E-03 & 1.29E-02 & \underline{1.11}
& \textbf{2.58E+04} & 3.05E+02 & 1.16
& \textbf{1.37E-03} & 3.76E-01 & 0.82
& 5.43E-03 & 1.28E-02 & 1.09
& \textbf{1.38E+00} & 5.22E+00 & 0.86 \\
\hline
BNN (QR) (1000)
& 6.42E+03 & 2.13E+02 & 0.98
& 3.75E-01 & 1.10E+00 & 0.72
& 2.51E+04 & 4.88E+02 & 1.08
& 7.06E-01 & 1.91E+00 & 0.23
& 5.77E-01 & 1.30E+00 & 0.55
& 1.18E+00 & 5.35E+00 & 0.85 \\
BNN + DR ($\tau=0.90$)
& \textbf{5.26E+03} & 2.18E+02 & \underline{0.98}
& \textbf{2.75E-01} & \textbf{1.07E+00} & \textbf{0.75}
& 3.13E+04 & 5.53E+02 & 1.06
& \textbf{6.67E-01} & \textbf{1.84E+00} & \textbf{0.27}
& \textbf{5.26E-01} & \textbf{1.29E+00} & \textbf{0.56}
& \textbf{9.09E-01} & 5.54E+00 & 0.83 \\
\hline
\multicolumn{19}{c}{} \\
\hline
Methods
& \multicolumn{3}{c}{DTLZ7}
& \multicolumn{3}{c}{Omnitest}
& \multicolumn{3}{c}{BNH}
& \multicolumn{3}{c}{Truss2D}
& \multicolumn{3}{c}{Welded Beam} \\
\cmidrule(lr){2-4} \cmidrule(lr){5-7} \cmidrule(lr){8-10} \cmidrule(lr){11-13} \cmidrule(lr){14-16}
& MSE & IGD+ & HV
& MSE & IGD+ & HV
& MSE & IGD+ & HV
& MSE & IGD+ & HV
& MSE & IGD+ & HV \\
\midrule
GPR (RBF) (1000)
& 2.19E-09 & 3.28E-03 & 1.11
& 6.55E-11 & 5.58E-03 & 1.16
& 5.24E-07 & 1.91E-01 & 1.05
& 2.49E+08 & 7.28E+04 & 0.14
& 1.04E+01 & 2.94E-01 & 1.13 \\
GPR + DR ($\tau=0.90$)
& 2.94E-09 & 5.75E-03 & \underline{1.11}
& \textbf{6.04E-11} & 5.76E-03 & \underline{1.16}
& \textbf{5.11E-07} & \underline{1.91E-01} & \underline{1.05}
& 2.91E+08 & \underline{7.28E+04} & \underline{0.14}
& 1.34E+01 & 3.65E-01 & \underline{1.13} \\
\hline
GPR (Matérn) (1000)
& 2.40E-09 & 3.96E-03 & 1.11
& 5.38E-09 & 5.84E-03 & 1.16
& 6.09E-07 & 1.90E-01 & 1.05
& 2.98E+11 & 5.61E+04 & 0.33
& 2.74E-03 & 2.96E-01 & 1.16 \\
GPR + DR ($\tau=0.90$)
& \underline{2.40E-09} & 4.75E-03 & \underline{1.11}
& 6.08E-09 & 5.98E-03 & \underline{1.16}
& \textbf{5.99E-07} & \underline{1.90E-01} & \underline{1.05}
& \underline{2.98E+11} & \textbf{2.38E+04} & \textbf{0.65}
& 3.51E-03 & 3.02E-01 & \textbf{1.17} \\
\hline
Autogluon (QR) (1000)
& 3.31E+00 & 1.05E-02 & 1.11
& 6.09E-03 & 1.21E-02 & 1.16
& 7.35E-01 & 2.09E-01 & 1.04
& 4.13E+07 & 5.09E-02 & 1.00
& 4.11E-01 & 2.58E-01 & 1.18 \\
QR + DR ($\tau=0.90$)
& \textbf{2.62E+00} & 1.74E-02 & \underline{1.11}
& \textbf{3.57E-03} & 1.42E-02 & \underline{1.16}
& \textbf{6.51E-01} & 2.44E-01 & \underline{1.04}
& \textbf{3.96E+06} & \textbf{1.05E-02} & \textbf{1.01}
& 1.64E+00 & \textbf{2.04E-01} & \underline{1.18} \\
\hline
BNN (QR) (1000)
& 4.15E+00 & 8.68E-02 & 1.10
& 2.65E-03 & 7.05E-03 & 1.16
& 2.32E+00 & 2.65E+01 & 0.54
& 1.62E+13 & 1.05E+06 & 0.58
& 1.36E+01 & 4.98E+01 & 0.41 \\
BNN + DR ($\tau=0.90$)
& \textbf{3.78E+00} & 1.08E-01 & \underline{1.10}
& 2.82E-03 & 8.06E-03 & \underline{1.16}
& 3.35E+00 & \underline{2.65E+01} & \underline{0.54}
& \underline{1.62E+13} & \underline{1.05E+06} & \underline{0.58}
& \textbf{9.01E+00} & \underline{4.98E+01} & \underline{0.41} \\
\hline
\end{tabular}
}
\caption{Ablation study of optimization performance comparison across 11 benchmark problems using surrogate models trained on 1,000 samples and combined with \gls{NSGA-II}. \textbf{Bold} indicates improvement over the baseline without DR, and \underline{underline} indicates equal performance.}
\label{tab:opt_1000}
\end{table*}

\subsection{Comparative Analysis against Baselines}
We utilize \gls{NSGA-II} as the optimization algorithm for the dual-ranking–based implementations; thus, this is omitted in the table. Table~\ref{tab:detail_ranks} reports the ranking results of all methods across eight unconstrained benchmark problems. Table~\ref{tab:opt_baseline_1000} reports experimental comparison results with baseline offline data-driven multi-objective optimization methods on eight unconstrained benchmark problems, where the surrogate models are trained on a dataset with 1,000 samples. Table~\ref{tab:detail_1000_ranks} presents the ranking results corresponding to the above comparison results.

\begin{table*}[!htbp]
\centering
\resizebox{\textwidth}{!}{
\begin{tabular}{l ccc ccc ccc ccc ccc ccc ccc ccc}
\toprule
Methods 
& \multicolumn{3}{c}{DTLZ1} 
& \multicolumn{3}{c}{DTLZ2} 
& \multicolumn{3}{c}{DTLZ3} 
& \multicolumn{3}{c}{DTLZ4} 
& \multicolumn{3}{c}{DTLZ5} 
& \multicolumn{3}{c}{DTLZ6} 
& \multicolumn{3}{c}{DTLZ7} 
& \multicolumn{3}{c}{Omnitest} \\
\cmidrule(lr){2-4} \cmidrule(lr){5-7} \cmidrule(lr){8-10} \cmidrule(lr){11-13} 
\cmidrule(lr){14-16} \cmidrule(lr){17-19} \cmidrule(lr){20-22} \cmidrule(lr){23-25}
& MSE & IGD+ & HV 
& MSE & IGD+ & HV 
& MSE & IGD+ & HV 
& MSE & IGD+ & HV 
& MSE & IGD+ & HV 
& MSE & IGD+ & HV 
& MSE & IGD+ & HV 
& MSE & IGD+ & HV \\
\midrule
Prob-RVEA        
& 8 & 8 & 8  & 3 & 4 & 4  & 8 & 8 & 8  & 3 & 2 & 2 
& 3 & 4 & 3  & 2 & 8 & 8  & 6 & 7 & 9  & 2 & 2 & 4 \\
Prob-MOEA/D      
& 9 & 9 & 9  & 1 & 5 & 4  & 9 & 9 & 9  & 1 & 7 & 7 
& 1 & 6 & 5  & 1 & 9 & 9  & 5 & 8 & 8  & 1 & 8 & 7 \\
IBEA-MS          
& 2 & 5 & 5  & 7 & 1 & 1  & 7 & 1 & 1  & 2 & 3 & 3 
& 7 & 2 & 1  & 8 & 1 & 1  & 3 & 9 & 1  & -- & -- & -- \\
TGPR-MO          
& 3 & 4 & 4  & 4 & 2 & 1  & 4 & 5 & 6  & 6 & 4 & 5 
& 4 & 1 & 1  & 7 & 2 & 2  & 2 & 2 & 1  & 3 & 5 & 6 \\
DDMOEA/GAN       
& 1 & 1 & 1  & 9 & 9 & 9  & 1 & 2 & 1  & 9 & 8 & 8 
& 9 & 9 & 9  & 9 & 6 & 6  & 9 & 6 & 7  & 7 & 6 & 2 \\
GPR(RBF)+DR      
& 7 & 2 & 2  & 6 & 7 & 7  & 3 & 3 & 3  & 4 & 6 & 4 
& 6 & 7 & 7  & 4 & 4 & 4  & 4 & 4 & 4  & 5 & 4 & 5 \\
GPR(Mat\'ern)+DR 
& 5 & 3 & 3  & 2 & 3 & 3  & 2 & 4 & 3  & 5 & 5 & 6 
& 2 & 3 & 3  & 6 & 3 & 3  & 1 & 1 & 1  & 6 & 3 & 2 \\
QR+DR            
& 4 & 6 & 6  & 5 & 6 & 6  & 6 & 6 & 5  & 7 & 1 & 1 
& 5 & 5 & 5  & 3 & 7 & 7  & 8 & 3 & 4  & 4 & 7 & 7 \\
BNN+DR           
& 6 & 7 & 7  & 8 & 8 & 8  & 5 & 7 & 6  & 8 & 9 & 9 
& 8 & 8 & 8  & 5 & 5 & 5  & 7 & 5 & 4  & 8 & 1 & 1 \\
\bottomrule
\end{tabular}
}
\caption{Ranking results of all methods across 8 unconstrained benchmark problems. Rank 1 indicates the best performance. The symbol ``-'' indicates the experiment could not be conducted.}
\label{tab:detail_ranks}
\end{table*}

\begin{table*}[htbp]
\centering
\resizebox{\textwidth}{!}{
\setlength{\tabcolsep}{3.5pt}
\begin{tabular}{l ccc ccc ccc ccc}
\toprule
Methods 
& \multicolumn{3}{c}{DTLZ1} 
& \multicolumn{3}{c}{DTLZ2} 
& \multicolumn{3}{c}{DTLZ3} 
& \multicolumn{3}{c}{DTLZ4} \\
\cmidrule(lr){2-4} \cmidrule(lr){5-7} \cmidrule(lr){8-10} \cmidrule(lr){11-13}
& MSE & IGD+ & HV
& MSE & IGD+ & HV
& MSE & IGD+ & HV
& MSE & IGD+ & HV \\
\midrule

Prob-RVEA
& 3.09E+04 & 1.95E+02 & 0.91
& 1.93E+00 & 9.84E-01 & 0.66
& 4.74E+05 & 7.40E+02 & 0.66
& 1.15E+00 & 9.84E-01 & 0.62 \\

Prob-MOEA/D
& 8.44E+03 & 1.97E+02 & 0.95
& 1.77E+00 & 1.06E+00 & 0.58
& 4.61E+05 & 6.87E+02 & 0.71
& 1.54E+00 & 8.51E-01 & 0.64 \\

IBEA-MS
& 4.67E+03  & 1.40E+02  & 1.13
& 5.06E-02  & 1.36E-02  & 1.11
& 3.74E+02  & 4.06E+02  & 1.14
& 5.24E-03  & 3.42E-01  & 0.98 \\

TGPR-MO
& 4.19E+03 & 1.99E+02 & 1.01
& 4.02E-04 & 2.06E-02 & 1.07
& 2.47E+04 & 5.11E+02 & 1.08
& 8.47E-01 & 8.21E-01 & 0.86 \\

DDMOEA/GAN
& 3.72E+03 & 9.18E+01 & 1.18
& 3.45E+00 & 1.54E+00 & 0.47
& 9.55E+03 & 2.26E+02 & 1.19
& 2.16E+00 & 9.98E-01 & 0.58 \\

GPR (RBF)+DR ($\tau=0.90$)
& 1.21E+04 & 1.34E+02 & 1.12
& 1.81E-08 & 2.58E-03 & 1.11
& 2.86E+04 & 4.05E+02 & 1.12
& 2.06E-01 & 2.76E-01 & 1.02 \\

GPR (Mat\'ern)+DR ($\tau=0.90$)
& 1.49E+04 & 1.32E+02 & 1.13
& 4.51E-05 & 3.09E-03 & 1.11
& 1.09E+05 & 3.23E+02 & 1.16
& 4.28E-02 & 1.09E-01 & 1.08 \\

QR+DR ($\tau=0.90$)
& 4.25E+03 & 1.34E+02 & 1.11
& 5.43E-03 & 1.29E-02 & 1.11
& 2.58E+04 & 3.05E+02 & 1.16
& 1.37E-03 & 3.76E-01 & 0.82 \\

BNN+DR ($\tau=0.90$)
& 5.26E+03 & 2.18E+02 & 0.98
& 2.75E-01 & 1.07E+00 & 0.75
& 3.13E+04 & 5.53E+02 & 1.06
& 6.67E-01 & 1.84E+00 & 0.27 \\

\midrule
Methods 
& \multicolumn{3}{c}{DTLZ5} 
& \multicolumn{3}{c}{DTLZ6} 
& \multicolumn{3}{c}{DTLZ7} 
& \multicolumn{3}{c}{Omnitest} \\
\cmidrule(lr){2-4} \cmidrule(lr){5-7} \cmidrule(lr){8-10} \cmidrule(lr){11-13}
& MSE & IGD+ & HV
& MSE & IGD+ & HV
& MSE & IGD+ & HV
& MSE & IGD+ & HV \\
\midrule

Prob-RVEA
& 1.82E+00 & 8.99E-01 & 0.65
& 4.08E+01 & 8.24E+00 & 0.22
& 8.45E+01 & 8.72E+00 & 0.78
& 1.16E-02 & 1.45E-01 & 1.13 \\

Prob-MOEA/D
& 1.45E+00 & 8.71E-01 & 0.60
& 4.16E+01 & 8.09E+00 & 0.27
& 8.40E+01 & 8.63E+00 & 0.81
& 2.82E-05 & 4.64E-01 & 1.02 \\

IBEA-MS
& 5.06E-02  & 1.26E-02  & 1.09
& 8.27E+00  & 1.23E+00  & 1.16
& 1.56E-02  & 1.89E+00  & 1.11
& - & - & - \\

TGPR-MO
& 3.87E-04 & 9.38E-03 & 1.10
& 2.13E+00 & 3.61E+00 & 1.00
& 2.38E-08 & 3.18E-01 & 1.09
& 1.16E-02 & 1.61E-02 & 1.16 \\

DDMOEA/GAN
& 3.19E+00 & 1.88E+00 & 0.17
& 4.97E+00 & 4.85E+00 & 0.90
& 9.91E+00 & 4.90E-02 & 1.10
& 1.61E+00 & 3.13E+00 & 0.20 \\

GPR (RBF)+DR ($\tau=0.90$)
& 1.81E-08 & 2.74E-03 & 1.10
& 9.51E-01 & 4.66E+00 & 0.92
& 2.94E-09 & 5.75E-03 & 1.11
& 6.04E-11 & 5.76E-03 & 1.16 \\

GPR (Mat\'ern)+DR ($\tau=0.90$)
& 4.51E-05 & 3.32E-03 & 1.10
& 2.45E+00 & 3.99E+00 & 0.99
& 2.40E-09 & 4.75E-03 & 1.11
& 6.08E-09 & 5.98E-03 & 1.16 \\

QR+DR ($\tau=0.90$)
& 5.43E-03 & 1.28E-02 & 1.09
& 1.38E+00 & 5.22E+00 & 0.86
& 2.62E+00 & 1.74E-02 & 1.11
& 3.57E-03 & 1.42E-02 & 1.16 \\

BNN+DR ($\tau=0.90$)
& 5.26E-01 & 1.29E+00 & 0.56
& 9.09E-01 & 5.54E+00 & 0.83
& 3.78E+00 & 1.08E-01 & 1.10
& 2.82E-03 & 8.06E-03 & 1.16 \\

\bottomrule
\end{tabular}
}
\caption{Optimization performance comparison with baseline offline data-driven multi-objective optimization methods on 8 unconstrained benchmark problems, where the surrogate models are trained on a dataset with 1,000 samples. Our method is combined with \gls{NSGA-II} (omitted in the table) and the dual-ranking (DR) strategy with $\tau = 0.9$. For each indicator, lower values indicate better performance for \gls{MSE} and \gls{IGD+}, while higher values indicate better performance for \gls{HV}. The symbol ``-'' indicates that the experiment cannot be conducted under that setting.}
\label{tab:opt_baseline_1000}
\end{table*}

\begin{table*}[!h]
\centering
\resizebox{\textwidth}{!}{
\begin{tabular}{l ccc ccc ccc ccc ccc ccc ccc ccc}
\toprule
Methods 
& \multicolumn{3}{c}{DTLZ1} 
& \multicolumn{3}{c}{DTLZ2} 
& \multicolumn{3}{c}{DTLZ3} 
& \multicolumn{3}{c}{DTLZ4} 
& \multicolumn{3}{c}{DTLZ5} 
& \multicolumn{3}{c}{DTLZ6} 
& \multicolumn{3}{c}{DTLZ7} 
& \multicolumn{3}{c}{Omnitest} \\
\cmidrule(lr){2-4} \cmidrule(lr){5-7} \cmidrule(lr){8-10} \cmidrule(lr){11-13} \cmidrule(lr){14-16} \cmidrule(lr){17-19} \cmidrule(lr){20-22} \cmidrule(lr){23-25}
& MSE & IGD+ & HV
& MSE & IGD+ & HV
& MSE & IGD+ & HV
& MSE & IGD+ & HV
& MSE & IGD+ & HV
& MSE & IGD+ & HV
& MSE & IGD+ & HV
& MSE & IGD+ & HV \\
\midrule
Prob-RVEA
& 9 & 6 & 9 & 8 & 6 & 7 & 9 & 9 & 9 & 7 & 7 & 7 & 8 & 7 & 6 & 8 & 9 & 9 & 9 & 9 & 9 & 6 & 6 & 6 \\
Prob-MOEA/D
& 6 & 7 & 8 & 7 & 7 & 8 & 8 & 8 & 8 & 8 & 6 & 6 & 7 & 6 & 7 & 9 & 8 & 8 & 8 & 8 & 8 & 3 & 7 & 7 \\
IBEA-MS
& 4 & 5 & 2 & 5 & 4 & 1 & 1 & 5 & 4 & 2 & 3 & 3 & 5 & 4 & 4 & 7 & 1 & 1 & 4 & 7 & 1 & - & - & - \\
TGPR-MO
& 2 & 8 & 6 & 3 & 5 & 5 & 3 & 6 & 6 & 6 & 5 & 4 & 3 & 3 & 1 & 4 & 2 & 2 & 3 & 6 & 7 & 6 & 5 & 1 \\
DDMOEA/GAN
& 1 & 1 & 1 & 9 & 9 & 9 & 2 & 1 & 1 & 9 & 8 & 8 & 9 & 9 & 9 & 6 & 5 & 5 & 7 & 4 & 5 & 8 & 8 & 8 \\
GPR (RBF)+DR
& 7 & 3 & 4 & 1 & 1 & 1 & 5 & 4 & 5 & 4 & 2 & 2 & 1 & 1 & 1 & 2 & 4 & 4 & 2 & 2 & 1 & 1 & 1 & 1 \\
GPR (Mat\'ern)+DR
& 8 & 2 & 2 & 2 & 2 & 1 & 7 & 3 & 2 & 3 & 1 & 1 & 2 & 2 & 1 & 5 & 3 & 3 & 1 & 1 & 1 & 2 & 2 & 1 \\
QR+DR
& 3 & 3 & 5 & 4 & 3 & 1 & 4 & 2 & 2 & 1 & 4 & 5 & 4 & 5 & 4 & 3 & 6 & 6 & 5 & 3 & 1 & 5 & 4 & 1 \\
BNN+DR
& 5 & 9 & 7 & 6 & 8 & 6 & 6 & 7 & 7 & 5 & 9 & 9 & 6 & 8 & 8 & 1 & 7 & 7 & 6 & 5 & 5 & 4 & 3 & 1 \\
\bottomrule
\end{tabular}
}
\caption{Ranking results of all methods across 8 unconstrained benchmark problems, where the surrogate models are trained on a dataset with 1,000 samples. Rank 1 indicates the best performance. The symbol ``-'' indicates the experiment could not be conducted.}
\label{tab:detail_1000_ranks}
\end{table*}


\bibliographystyle{named}
\bibliography{ijcai26}